\documentclass[10pt,journal,compsoc]{IEEEtran}
\usepackage{amssymb}

\ifCLASSOPTIONcompsoc
  \usepackage[nocompress]{cite}
\else
  \usepackage{cite}
\fi

\ifCLASSINFOpdf
  \usepackage[pdftex]{graphicx}
\else
\fi

\usepackage{algorithm}
\usepackage{algpseudocode}

\usepackage{array}

\ifCLASSOPTIONcompsoc
 \usepackage[caption=false,font=footnotesize,labelfont=sf,textfont=sf]{subfig}
\else
 \usepackage[caption=false,font=footnotesize]{subfig}
\fi
\usepackage{dblfloatfix}

\ifCLASSOPTIONcaptionsoff
 \usepackage[nomarkers]{endfloat}
\let\MYoriglatexcaption\caption
\renewcommand{\caption}[2][\relax]{\MYoriglatexcaption[#2]{#2}}
\fi
\usepackage{url}

\usepackage{amsmath,amsthm}
\usepackage{float,xspace}
\usepackage{thm-restate}
\usepackage{balance}
\usepackage{cite,enumerate}
\usepackage{footnote}
\usepackage{float}

\usepackage{pifont}
\newcommand{\cmark}{\ding{51}}%
\newcommand{\xmark}{\ding{55}}%
\usepackage{pbox}
\usepackage{cellspace}
\usepackage[flushleft]{threeparttable}
\usepackage[table]{xcolor}

\usepackage[keeplastbox]{flushend}
\usepackage{placeins}

\usepackage{hyperref}
\usepackage{framed}

\newcommand{\E}{\mathbb{E}}

\usepackage{bbm}
\usepackage{verbatim}

\newcommand\numberthis{\addtocounter{equation}{1}\tag{\theequation}}

\newcommand{\dist}{\mathrm{dist}}

\newcommand{\RMSE}{\mathrm{RMSE}}

\newcommand{\R}{\mathbb{R}}

 \newtheorem{thm}{Theorem}

\newtheorem{lem}[thm]{Lemma}
 \newtheorem{prop}[thm]{Proposition}
\newtheorem{defi}[thm]{Definition}
\newtheorem{obs}[thm]{Observation}

\newcommand{\fsketch}{{\tt FSketch}\xspace}
\newcommand{\minfsketch}{{\tt Median-FSketch}\xspace}

\newcommand{\bl}[1]{{#1}}

\usepackage[utf8]{inputenc} 
\usepackage{hyperref}       
\usepackage{url}            
\usepackage{booktabs}       
\usepackage{nicefrac}       
\usepackage{paralist}
\usepackage{todonotes}

\begin{document}
%
\title{Dimensionality Reduction for Categorical Data}
%
%
%
%

\author{Debajyoti~Bera,
        Rameshwar~Pratap,
        and~Bhisham~Dev~Verma
\IEEEcompsocitemizethanks{\IEEEcompsocthanksitem D. Bera is with the Department
of Computer Science and Engineering, Indraprastha Institute of Information Technology (IIIT-Delhi), New Delhi, India, 110020.\protect\\
E-mail: see http://www.michaelshell.org/contact.html
\IEEEcompsocthanksitem R. Pratap and B. D. Verma are with the Indian Institute of Technology, Mandi, Himachal Pradesh, India.\protect\\
\IEEEcompsocthanksitem Emails: dbera@iiitd.ac.in, rameshwar@iitmandi.ac.in and d18039@students.iitmandi.ac.in.}
\thanks{Manuscript accepted for publication by IEEE Transactions on Knowledge and Data Engineering. Copyright 1969, IEEE.}}

%
%

\markboth{Journal of \LaTeX\ Class Files,~Vol.~14, No.~8, August~2015}%
{Shell \MakeLowercase{\textit{et al.}}: Bare Demo of IEEEtran.cls for Computer Society Journals}
%



\IEEEtitleabstractindextext{%
\begin{abstract}
Categorical attributes are those that can take a discrete set of values, e.g., colours. This work is about compressing vectors over categorical attributes to low-dimension discrete vectors. The current hash-based methods compressing vectors over categorical attributes to low-dimension discrete vectors do not provide any guarantee on the Hamming distances between the compressed representations. Here we present \fsketch to create sketches for sparse categorical data and an estimator to estimate the pairwise Hamming distances among the uncompressed data only from their sketches. We claim that these sketches can be used in the usual data mining tasks in place of the original data without compromising the quality of the task. For that, we ensure that the sketches also are categorical, sparse, and the Hamming distance estimates are reasonably precise. Both the sketch construction and the Hamming distance estimation algorithms require just a single-pass; furthermore, changes to a data point can be incorporated into its sketch in an efficient manner. The compressibility depends upon how sparse the data is and is independent of the original dimension -- making our algorithm attractive for many real-life scenarios. Our claims are backed by rigorous theoretical analysis of the properties of \fsketch and supplemented by extensive comparative evaluations with related algorithms on some real-world datasets. We show that \fsketch is significantly faster, and the accuracy obtained by using its sketches are among the top for the standard unsupervised tasks of $\mathrm{RMSE}$, clustering and similarity search.

\end{abstract}

}

\maketitle

\IEEEdisplaynontitleabstractindextext

%
\IEEEpeerreviewmaketitle


\section{Introduction}
Of the many types of digital data that are getting recorded every second, 
most can be ordered -- they belong to the ordinal type (e.g., age, citation count, {\it etc.}), and a good proportion can be represented as strings but cannot be ordered --- they belong to the nominal type (e.g., hair colour, country, publication venue, {\it etc.}). The latter datatype is also known as {\em categorical} which is our focus in this work. Categorical attributes are commonly present in survey responses, and have been used earlier to model problems in bio-informatics~\cite{DNA,HIV}, market-basket transactions~\cite{transaction,itemset_categorical,itemset_mining}, web-traffic~\cite{web_transaction}, images~\cite{image_categorical}, and recommendation systems~\cite{click_stream}. The first challenge practitioners encounter with such data is how to process them using standard tools most of which are designed for numeric data, that too often are real-valued.

Two important operations are often performed before running statistical data analysis tools and machine learning algorithms on such datasets. \bl{The first is encoding the data points using numbers, and the second is dimensionality reduction; many approaches combine the two, with the final objective being numeric vectors of fewer dimensions. To the best of our knowledge, the approaches usually followed are ad-hoc adaptations of those employed for vectors in the real space, and suffer from computational inefficiency and/or unproven heuristics~\cite{Hancock2020SurveyOC}. The motivation of this work is to provide a solution that is efficient in practice and has proven theoretical guarantees.}

For the first operation, we use the standard method of label encoding in this paper. In this a feature with $c$ categories is represented by an integer from $\{0,1,2,\ldots c\}$ where 0 indicates a missing \bl{category} and $i \in \{1,2,\ldots, c\}$ indicates the $i$-th \bl{category}. Hence, an $n$-dimensional data point, where each feature can take at most $c$ values, can be represented by a vector from $\{0,1,2\ldots c\}^n$ --- we call such a vector as a categorical vector. Another approach is one-hot encoding (OHE) which is more popular since it avoids the implicit ordering among the feature values imposed by label-encoding. One-hot encoding of a feature with $c$ possible values is a $c$-dimensional binary vector in which the $i$-th bit is set to 1 to represent the $i$-th feature value. Naturally, one-hot encoding of an $n$-dimensional vector will be $nc$ dimensional --- which can be very large if $c$ is large (e.g., for features representing countries, {\it etc.}). Not only label encoding avoids this problem, but is essential for the crucial second step -- that of dimensionality reduction.

Dimensionality reduction is important when data points lie in a high-dimensional space, e.g., when encoded using one-hot encoding or when described using tens of thousands of categorical attributes. High-dimensional data vectors not only increase storage and processing cost, but they suffer from the ``curse of dimensionality'' that points to the decrease in performance after the dimension of the data points crosses a peak. Hence it is suggested that the high-dimensional categorical vectors be compressed to smaller vectors, essentially retaining the information only from the useful features. Baraniuk et al.~\cite{baraniuk2010low} characterised a good dimensionality reduction {\em in the Euclidean space} as a compression algorithm that satisfies the following two conditions for any two vectors $x$ and $y$.

\begin{enumerate}
    \item Information preserving: For any two distinct vectors $x$ and $y$, $R(x) \not= R(y)$.
    \item $\epsilon$-Stability: (Euclidean) distances between all the points are approximately preserved (with $\epsilon$ inaccuracy).
\end{enumerate}

We call these two conditions the {\em ``well-designed''} conditions. To obtain their mathematically precise versions, we need to narrow down upon a distance measure for categorical vectors. A natural measure for categorical vectors is an extension of the binary Hamming distance. For two $n$-dimensional categorical data points $x$ and $y$, the Hamming distance between them is defined as the number of features with different attributes in $x$ and $y$, i.e., 
\begin{align*}
HD(x, y) &=\Sigma_{i=1}^n \dist(x[i], y[i]) \mbox{, where }\\
\dist(x[i], y[i])&=\begin{cases}
    1, & \text{if~} x[i]\neq y[i], \\
    0, & \text{otherwise}.
  \end{cases}
\end{align*}

{\bf Problem statement:} The specific problem that we address is how to design a dimensionality reduction algorithm that can compress high-dimensional sparse label-encoded categorical vectors to low-dimensional categorical vectors so that (a) compressions of distinct vectors are distinct, and (b) the Hamming distance between two uncompressed vectors can be efficiently approximated from their compressed forms. These conditions, in turn, guarantee both information-preserving and stability. Furthermore, we would like to take advantage of the sparse nature of many real-world datasets. \bl{The} most important requirement is the compressed vectors should be categorical as well, specifically not over real numbers and preferably not binary; this is to allow the statistical tests and machine learning tools for categorical datasets, e.g. k-mode, to run on the compressed datasets.

\subsection{\bl{Challenges in the existing approaches}} Dimensionality reduction is a well-studied problem~\cite{plos_categorical_tips} (also see Table~\ref{table:dim-red} in Appendix) but Hamming space does not allow the usual approaches applicable in the Euclidean spaces. Methods that work for continuous-valued data or even ordinal data (such as integers) do not perform satisfactorily for unordered categorical data. Among those that specifically consume categorical data, techniques \textit{via} feature selection \bl{have} been well studied. For example, in the case of  labelled data $\chi^2$~\cite{chi_square} and {Mutual Information}~\cite{MI} based methods select features based on their correlation with the label. This limits their applicability to only the classification tasks. Further,  {Kendall rank correlation coefficient}~\cite{kendall1938measure} ``learns'' the important features based on the correlation among them. Learning approaches tend to be computationally heavy and do not work reliably with small training samples. So what about task-agnostic approaches that do not involve learning? PCA-based methods, e.g., MCA is popular among the practitioners of biology~\cite{plos_categorical_tips}; however, we consider them merely a better-than-nothing approach since PCA is fundamentally designed for continuous data.

\begin{figure}[!ht]
    \includegraphics[width=0.6\linewidth]{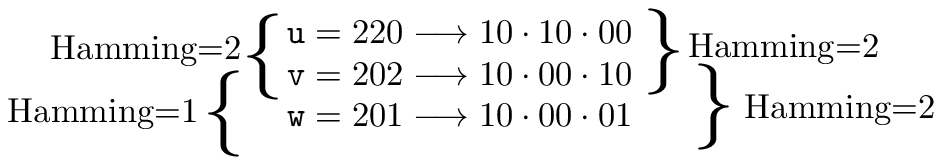}
    \caption{An example showing that the Hamming distances of one-hot encoded sparse vectors are not functionally related to the distances between their unencoded forms. \bl{If a feature, say country, is missing, libraries differ in their handling of its one-hot encoding. In this paper, we follow the common practice of using the $c$-dimensional all-zero vector as its encoding. This retains sparsity since the number of non-missing attributes in the original vector equals the number of non-zero bits in the encoded vector.} \label{fig:ohe-counterexample}}
\end{figure}

A quick search among internet forums, tutorials and Q\&A websites revealed that the more favourable approach to perform machine learning tasks on categorical datasets is to convert categorical feature vectors to binary vectors using one-hot encoding~\cite[see DictVectorizer]{scikit-learn} --- a widely-viewed tutorial on Kaggle calls it ``The Standard Approach for Categorical Data''~\cite{kaggle_one_hot}. \bl{The biggest problem with OHE is that it is impractical for large $n$ or large $c$ followed by a technical annoyance that some OHE implementations do not preserve the Hamming distances for sparse vectors} (see illustration in Figure~\ref{fig:ohe-counterexample}). Hence, this encoding is used in conjunction with problem-specific feature selection or followed by dimensionality reduction from binary to binary vectors~\cite{ICDM,oddsketch,JS_BCS}. The latter is a viable heuristic that we wanted to improve upon by allowing non-binary compressed vectors (see Appendix~\ref{appendix:OHE+BS} for a quick analysis of OHE followed by a state-of-the-art binary compression).

Another popular alternative, especially when $n\times c$ is large, is {\em feature hashing}~\cite{WeinbergerDLSA09} that is now part of most libraries, e.g., {\tt scikit-learn}~\cite[see FeatureHasher]{scikit-learn}. Feature hashing and other forms of hash-based approaches, also known as sketching algorithms, both encode and compress categorical feature vectors into integer vectors (sometimes signed) of a lower dimension, and furthermore, provide theoretical guarantees like stability, in some metric space. The currently known results for feature hashing apply only to the Euclidean space, however, Euclidean distance and Hamming distance are not monotonic for categorical vectors. It is neither known nor straightforward to ascertain whether feature hashing and its derivatives can be extended to the Hamming space which lacks the continuity that is crucial to their theoretical bounds. Other hash-based approaches either come with no guarantees and are used merely because of their compressibility or come with stability-like guarantees in a different space, e.g., cosine similarity by Simhash~\cite{simhash}. Our solution is a hashing approach that we prove to be stable in the Hamming space.


\subsection{Overview of results} 
The commonly followed practices in dealing with categorical vectors, especially those with high dimensions and not involving supervised learning or training data, appear to be either feature hashing or one-hot encoding followed by dimensionality reduction of binary vectors~\cite[Chapter 5]{zheng2018feature}. We provide a contender to these in the form of the {\em \fsketch sketching algorithm} to construct lower-dimensional categorical vectors from high-dimensional ones.

The lower-dimensional vectors, {\it sketches}, produced by \fsketch (we shall call these vectors as \fsketch too) have the desired theoretical guarantees and perform well on real-world datasets vis-\`a-vis related algorithms. Now we summarise the important features of \fsketch; in the summarisation, $p$ is a constant that is typically chosen to be a prime number between 5-50.

{\bf Lightweight and unsupervised:} First and foremost, \fsketch is an unsupervised process, and in fact, quite lightweight making a single pass over an input vector and taking $O(poly(\log p))$ steps per non-missing feature. The \fsketch-es retain the sparsity of the input vectors and their size and dimension do not depend at all on $c$. To make our sketches applicable out-of-the-box for modern applications where data keeps changing, we present an extremely lightweight algorithm to incorporate any change in a feature vector into its sketch in $O(poly(\log p))$-steps per modified feature. It should be noted that \fsketch supports change of an attribute, deletion of an attribute and insertion of a previously missing attribute unlike some state-of-the-art sketches; for example,  BinSketch~\cite{ICDM} does not support deletion of an attribute.

{\bf Estimator for Hamming distance:} We want to advocate the use of \fsketch-es for data analytic tasks like clustering, {\it etc.} that use Hamming distance for the (dis)similarity metric. We present an {\em estimator that can approximate the Hamming distance} between two points by making a single pass over their sketches. The estimator follows a tight concentration bound and 
has the ability to estimate the Hamming distance from very low-dimensional sketches. In the theoretical bounds, the dimensions could go as low as $4\sigma$ or even $\sqrt{\sigma}$ (and independent of the dimension of the data) where $\sigma$ indicates the sparsity (maximum number of non-zero attributes) of the input vectors; however, we later show that a much smaller dimension suffices in practice. Our sketch generation and the Hamming distance estimation algorithms combined meet the two conditions of ``well-designed'' dimensionality reduction.

\begin{thm} \label{thm:main-intro} Let $x$ and $y$ be distinct categorical vectors, and $\phi(x)$ and $\phi(y)$ be their $d$-dimensional compressions.
\begin{enumerate}
    \item $\phi(x)$ and $\phi(y)$ are distinct with probability $\approx HD(x,y)/d$.
    \item Let $HD'(x,y)$ denote the approximation to the Hamming distance between $x$ and $y$ computed from $\phi(x)$ and $\phi(y)$. If $d$ is set to $4\sigma$, then with probability at least $1-\delta$ (for any $\delta$ of choice), $$\big|HD(x,y) - HD'(x,y)\big| = { O \left( \sqrt{\sigma \ln \tfrac{2}{\delta}}\right)}.$$
\end{enumerate}
\end{thm}

The proof of (1) follows from Lemma~\ref{lem:expectation-previous} and the proof of (2) follows from Lemma~\ref{lem:hconcentrationlemma} for which we used McDiarmid's inequality. The theorem allows us to use compressed forms of the vectors in place of their original forms for data analytic and statistical tools that depend largely on their pairwise Hamming distances.


{\bf Practical performance:} All of the above claims are proved rigorously but one may wonder how do they perform in practice.
For this, we design an elaborate array of experiments on real-life datasets involving many common approaches for categorical vectors. The experiments demonstrate these facts.
\begin{itemize}
    \item Some of the baselines do not output categorical vectors (see Section~\ref{sec:experiments}).   
    Our \fsketch algorithm is super-fast  among those that do and offer comparable accuracy.
    \item When used for typical data analytic tasks like clustering, similarity search, etc. low-dimension \fsketch-es bring immense speedup {\it vis-a-vis} using the original (uncompressed) vectors, yet achieving very high accuracy. The NYTimes dataset saw 140x speedup upon compression to $0.1\%$.
    \item Even though highly compressed, the results of clustering, {\it etc.}\ on \fsketch-es are close to what could be obtained from the uncompressed vectors and are comparable with the best alternatives. For example, we were able to compress the Brain cell dataset of dimensionality $1306127$ to $1000$ dimensions in a few seconds, yet retaining the ability to correctly approximating the pairwise Hamming distances from the compressed vectors. This is despite many other baselines giving either an out-of-memory error, not stopping even after running for a sufficiently long time, or producing significantly worse estimates of pairwise Hamming distances.
    \item The parameter $p$ can be used to fine-tune the quality of results and the storage of the sketches.
\end{itemize}

We claim that \fsketch is the best method today to compress categorical datasets for data analytic tasks that require pairwise Hamming distances with respect to both theoretical guarantee and practical performance.

\subsection{Organisation of the paper}
The rest of the paper is organised as follows. We discuss several related works in Section~\ref{sec:relwork}.
In Section~\ref{sec:analysis}, we present our algorithm \fsketch and derive its theoretical bounds. In Section~\ref{sec:experiments}, we empirically compare the performance of \fsketch on several end tasks with state-of-the-art algorithms. We conclude our presentation in Section~\ref{sec:conclusion}.
The proofs of the theoretical claims and the results of additional experiments are included in Appendix.

\section{Related work} \label{sec:relwork}
{\bf Dimensionality reduction:} 
Dimensionality reduction has been studied in-depth for real-valued vectors, and to some extent, also for discrete vectors. We categorise them into these broad categories --- (a) random projection, (b) spectral projection, (c) locality sensitive hashing (LSH), (d) other hashing approaches, and (e) learning-based algorithms. All of them compress high-dimensional input vectors to low-dimensional ones that explicitly or implicitly preserve some measure of similarity between the input vectors.

The seminal result by Johnson and Lindenstrauss~\cite{JL83} is probably the most well known random projection-based algorithm for dimensionality reduction. This algorithm compresses real-valued vectors to low-dimensional real-valued vectors such that the Euclidean distances between the pairs of vectors are approximately  preserved, but in such a manner that the compressed dimension does not depend upon the original dimension. The algorithm involves projecting a data matrix onto a random matrix whose each entry is sampled from a Gaussian distribution. This result has seen lots of enhancements, particularly with respect to generating the random matrix without affecting the accuracy~\cite{Achlioptas03}, \cite{LiHC06}, \cite{KaneN14}. However, it is not clear whether any of those ideas can be made to work for categorical data and that too, for approximating Hamming distances.


Principal component analysis (PCA) is a spectral projection-based technique for reducing the dimensionality of high dimensional datasets by creating new uncorrelated variables that successively maximise variance. There are extensions of PCA that employ kernel methods that try to capture non‐linear relationships~\cite{ScholkopfSM97}. {Multiple Correspondence Analysis (MCA)}~\cite{MCA} does the analogous job for the categorical datasets. However, these methods perform dimensionality reduction by creating un-correlated features in a low-dimensional space whereas our aim is to preserve the pairwise Hamming distances in a low-dimensional space.

Another line of dimensionality reduction techniques builds upon the ``Locality Sensitive Hashing (LSH)'' algorithms. LSH algorithms have been proposed for different data types and similarity measures, e.g., real-valued vectors and the Euclidean distance~\cite{IM98}, real-valued vectors and the cosine similarity~\cite{simhash}, binary vectors and the Jaccard similarity~\cite{BroderCFM98}, binary vectors and the Hamming distance~\cite{GIM99}. However, generally speaking, the objective of an LSH is to group items so that similar items are grouped together and dissimilar items are not; unlike \fsketch they do not provide explicit estimators of any similarity metric.


There are quite a few learning-based dimensionality reduction algorithms available such as {Latent Semantic Analysis (LSA)}\cite{LSI}, {Latent Dirichlet Allocation (LDA)}\cite{LDA}, {Non-negative Matrix Factorisation (NNMF)}\cite{NNMF}, \bl{{Generalized feature embedding learning (GEL)}~\cite{golinko2019generalized}} all of which strive to learn a low-dimensional representation of a dataset while preserving some inherent properties of the full-dimensional dataset. \bl{They are rather slow due to the optimization step involved during learning.} T-distributed Stochastic Neighbour Embedding \texttt{(t-SNE)}~\cite{vanDerMaaten2008} is a faster non-linear dimensionality reduction technique that is widely used for the visualisation of high-dimensional datasets. However, the low-dimensional representation obtained from \texttt{t-SNE} is not recommended for use for other end tasks such as clustering, classification, anomaly detection as it does not necessarily preserve densities or pairwise distances. An autoencoder~\cite{Goodfellow-et-al-2016} is another learning-based non-linear dimension reduction algorithm. It basically consists of two parts: An \textit{encoder} which aims to learn a low-dimensional representation of the input and a \textit{decoder} which tries to reconstruct the original input from the output of the encoder. However, these approaches involve optimising a learning objective function and are usually slow and CPU-intensive.

The other hashing approaches randomly assign each feature (dimension) to one of several bins, and then compute a summary value for each bin by aggregating all the feature values assigned to it. A list of such summaries can be viewed as a low-dimensional sketch of the input. Such techniques have been designed for real-valued vectors approximating inner product (e.g., feature hashing~\cite{WeinbergerDLSA09}), binary vectors allowing estimation of several similarity measures such as Hamming distance, Inner product, Cosine, and Jaccard similarity (e.g., BinSketch~\cite{ICDM}), etc. This work is similar to these approaches but for categorical vectors and only aiming to estimate the Hamming distances.

Another approach in this direction could be to encode categorical vectors to binary and then apply dimensionality reduction for binary vectors; unfortunately, the popular encodings, e.g. OHE, do not preserve Hamming distance for vectors with missing features. Nevertheless, it is possible to encode using OHE and then reduce its dimension. However, our theoretical analysis led to a worse accuracy compared to that of \fsketch (see Appendix~\ref{appendix:OHE+BS} for the analysis) and this approach turned out to be one of the worst performers in our experiments (see Section~\ref{sec:experiments}).

\bl{While our motivation was to design an end-task agnostic dimensionality reduction algorithm, there exist several that are designed for specific tasks, e.g., for clustering~\cite{8809826}, for regression and discriminant analysis of labelled data~\cite{JMLR:v21:17-788}, and for estimating covariance matrix~\cite{DBLP:journals/tnn/ChenYZLK20}. Deep learning has gained mainstream importance and several researchers have proposed a dimensionality reduction “layer” inside a neural network~\cite{8974600}; this layer is intricately interwoven with the other layers and cannot be separated out as a standalone technique that outputs compressed vectors.}

\bl{Feature selection is a limited form of dimensionality reduction whose task is to identify a set of good features, and maybe learn their relative importance too. Banerjee and Pal~\cite{7155531} recently proposed an unsupervised technique that identifies redundant features and selects those with bounded correlation, but only for real-valued vectors. For our experiments we chose the Kendall-Tau rank correlation approach that is applicable to discrete-valued vectors.}

\textbf{Sketching algorithm:} The use of ``sketches'' for computing Hamming distance has been explicitly studied in the streaming algorithm framework. The first well-known solution was proposed by Cormode et al.~\cite{CormodeDIM03} where they showed how to estimate a Hamming distance with high accuracy and low error. There have been several improvements to this result, in particular, by Kane et al.~\cite{KaneNW10} where a sketch with the optimal size was proposed. However, we neither found any implementation nor an empirical evaluation of those approaches (the algorithms themselves appear fairly involved). Further, their objective was to minimise the space usage in the asymptotic sense in a streaming setting, whereas, our objective is to design a solution that can be readily used for data analysis. This motivated us to compress categorical vectors onto low-dimensional categorical vectors, unlike the real-valued vectors that the theoretical results proposed. 
A downside of our solution is that it heavily relies on the sparsity of a dataset unlike the sketches output by the streaming algorithms.

\begin{table}[h]
    \centering
    \caption{Notations\label{tab:notations}}    
\noindent\begin{tabular}{lc}
    \hline
    categorical data vectors & $x,y$ \\
    their Hamming distance & $h$ \\
    \hline
    compressed categorical vectors (sketches) & $\phi(x), \phi(y)$ \\
    $j$-th bit of a sketch $\phi(x)$ & $\phi_j(x)$ \\
    observed Hamming distance between sketches & $f$ \\
    \hline
    expected Hamming distance between sketches & $f^*$\\
    estimated Hamming distance between data vectors & $\hat{h}$\\
    \hline
    \end{tabular}
\end{table}

\section{Category sketching and Hamming distance estimation}\label{sec:analysis}

Our technical objective is to design an effective algorithm to compress high-dimensional vectors over $\{0,1,\ldots, c\}$ to integer vectors of a low dimension, {\it aka.} sketches;  \bl{$c$ can even be set to an upper bound on the largest number of categories among all the features}. The number of attributes in the input vectors is denoted $n$ and the dimension of the compressed vector is denoted $d$. We will later show how to choose $d$ depending on the sparsity of a dataset that we denote $\sigma$. The commonly used notations in this section are listed in Table~\ref{tab:notations}.


    \begin{algorithm}
    \begin{algorithmic}[1]
    \Procedure{Initialize}{}
        \State Choose random mapping $\rho:\{1, \ldots n\} \to \{1, \ldots d\}$
        \State Choose some prime $p$
        \State Choose $n$ random numbers $R=r_1, \ldots, r_n$ with each $r_i \in \{0, \ldots p-1\}$
    \EndProcedure
    \end{algorithmic}
	\begin{algorithmic}[1]
	\Procedure{CreateSketch}{$x \in \{0,1,\ldots c\}^n$}
        \State Create empty sketch $\phi(x) = 0^d$
        \For{$i=1 \ldots n$}
            \State $j=\rho(i)$
            \State $\phi_j(x) = (\phi_j(x) + x_i \cdot r_i) \mod{p}$
        \EndFor
        \State \Return $\phi(x)$
	\EndProcedure
	\end{algorithmic}
	\caption{Constructing $d$-dimensional \fsketch of $n$-dimensional vector $x$\label{algo:fsketchalgo}}
    \end{algorithm}
    
\begin{figure}[t]
    \centering
    \includegraphics[width=\linewidth]{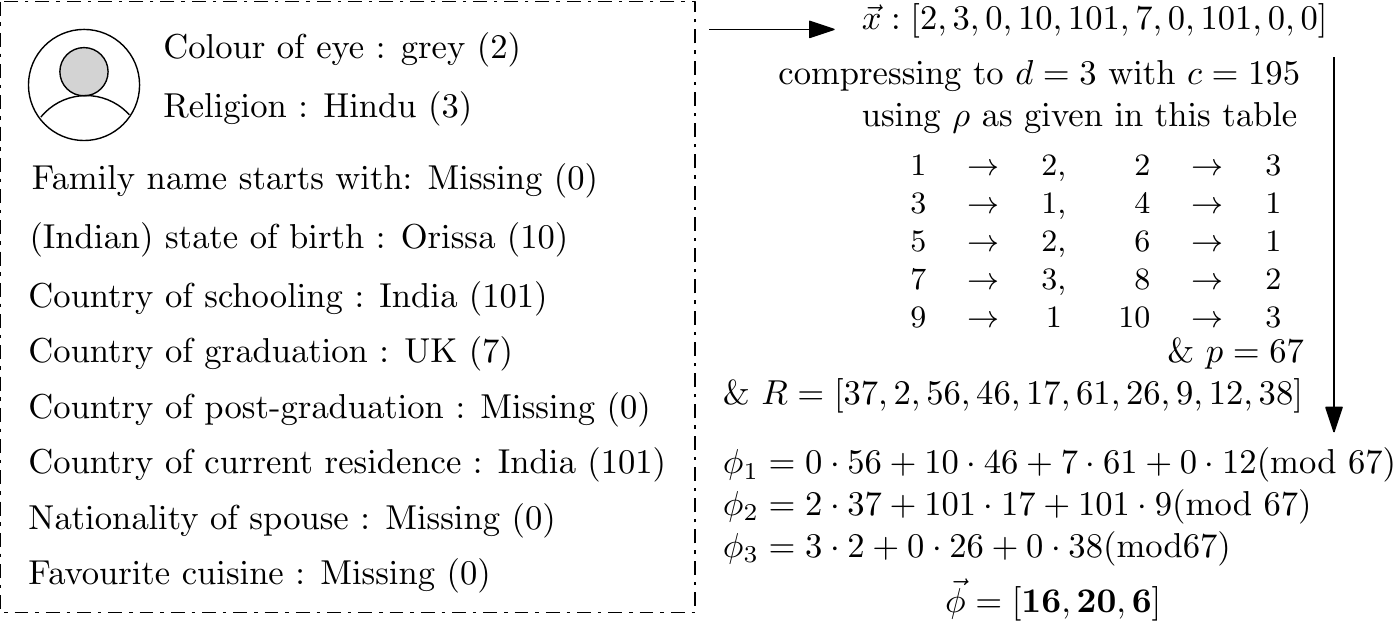}
    \caption{\bl{An example illustrating how to compress a data point with categorical features using \fsketch to a 3-dimensional integer vector. The data point has 10 feature values, each of which is a categorical variable (the corresponding label encoded values are present inside the brackets). $c$ is chosen as 195 since the fifth, sixth, seventh, and eighth features have 195 categories which is the largest. $\rho, p$ and $R$ are internal variables of \fsketch.}}
    \label{fig:example}
\end{figure}

\subsection{\fsketch construction}
Our primary tool for sketching categorical data is a randomised sketching algorithm named \fsketch that is described in Algorithm~\ref{algo:fsketchalgo}; see Figure~\ref{fig:example} for an example.

Let  $x \in \{0,1,\ldots c\}^n$ denote the input vector, and the $i$-th feature or co-ordinate of $x$ is denoted by $x_i$. The sketch of input vector $x$ will be denoted $\phi(x) \in \{0, 1, \ldots p-1\}^d$ whose coordinates will be denoted $\phi_1(x), \phi_2(x), \ldots, \phi_d(x)$. Note that the initialisation step of \fsketch needs to run only once for a dataset. We are going to use the following characterisation of the sketches in the rest of this section; \bl{a careful reader may observe the similarity to Freivald's algorithm for verifying matrix multiplication~\cite{Freivalds1977ProbabilisticMC}.}

\begin{obs}\label{obs:formula}
It is obvious from Algorithm~\ref{algo:fsketchalgo} that the sketches created by \fsketch satisfy $\phi_j(x) = (\sum_{i \in \rho^{-1}(j)} x_i \cdot r_i) \mod{p}$.
\end{obs}

\begin{figure*}
    \centering
    \includegraphics[width=\linewidth]{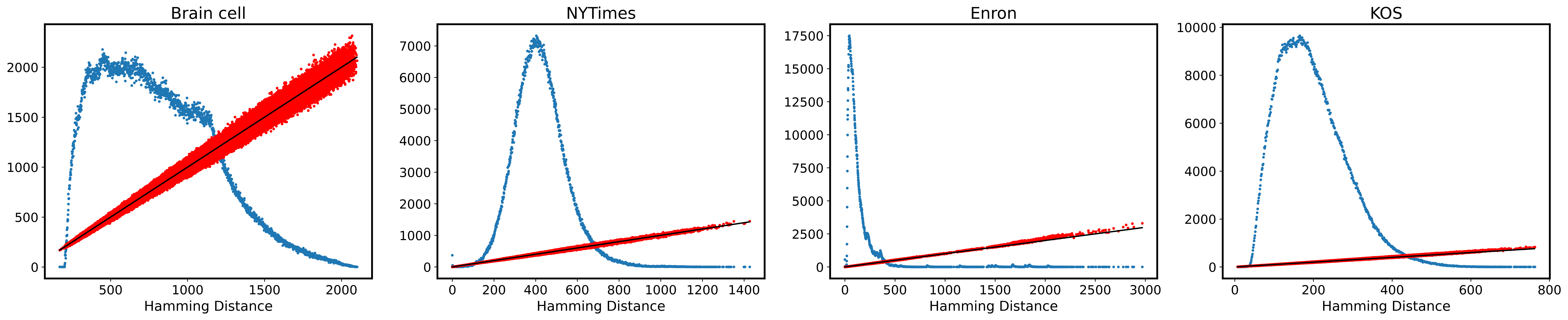}
    \caption{The distributions of Hamming distances for some of the datasets used in our experiments are shown in blue --- the Y-axis shows the frequency of each distance. The black points represent the actual Hamming distances and the red points are the estimates, i.e., a red-point plotted against a Hamming distance $d$ (on the X-axis) shows the estimated Hamming distance between two points with actual Hamming distance $d$. Observe that the Hamming distances follow a long-tailed distribution and that most distances are fairly low --- moreover, our estimates are more accurate for those high frequent Hamming distances.
    \label{fig:hamming_distance_distrib}}
\end{figure*}

\subsection{Hamming distance estimation}
Here we explain how the Hamming distance between $x$ and $y$ denoted $HD(x,y)$, percolates to their sketches as well. \bl{The objective is derive an estimator for $HD(x,y)$ from the Hamming distance between $\phi(x)$ and $\phi(y)$.}

The sparsity of a set of vectors denoted $\sigma$, is the maximum number of non-zero coordinates in them. For the theoretical analysis, we assume that we know the sparsity of the dataset, or at least an upper bound of the same. Note that, for a pair of sparse vectors $x, y \in \{0, 1, \ldots, c\}^n$, the  Hamming distance between them can vary from $0$ (when they are same) to $2\sigma$ (when they are completely different).

We first prove case (a) of Theorem~\ref{thm:main-intro} which states that sketches of different vectors are rarely the same.

\begin{lem}\label{lem:expectation-previous}
Let $h$ denote $HD(x,y)$ for two input vectors $x,y$ to \fsketch. Then
\begin{align*}
\Pr_{\rho,R} [\phi_j(x)\not=\phi_j(y)] &= (1-\tfrac{1}{p})(1-(1-\tfrac{1}{d})^h).
\end{align*}
\end{lem}


\begin{proof} 
    Fix a mapping $\rho$ and then define $F_j(x)$ as the vector $[x_{i_1}, x_{i_2}, \ldots ~:~ i_k \in \{1, \ldots n\}]$ of values of $x$ that are mapped to $j$ in $\phi(x)$ in the increasing order of their coordinates, i.e., $\rho(i_k)=j$ and $i_1 < \ldots i_k < i_{k+1}$. Since $\rho$ is fixed, $F_j(y)$ is also a vector of the same length. The key observation is that if $F_j(x)=F_j(y)$ then $\phi_j(x)=\phi_j(y)$ but the converse is not always true. Therefore we separately analyse both the conditions (a) $F_j(x)\not=F_j(y)$ and (b) $F_j(x)=F_j(y)$.

It is given that $x$ and $y$ differ at $h$ coordinates. Therefore, $F_j(x)\not= F_j(y)$ iff any of those coordinates are mapped to $j$ by $\rho$. Thus, 
\begin{align*}
    \Pr_\rho[F_j(x)=F_j(y)] = (1-\tfrac{1}{d})^h. \numberthis \label{eq:prob_pi}
\end{align*}

Next we analyse the chance of $\phi_j(x)=\phi_j(y)$ when $F_j(x)\not=F_j(y)$. Note that $\phi_j(x)=(x_{i_1} \cdot r_{i_1} + x_{i_2} \cdot r_{i_2} + \ldots) \mod{p}$ (and a similar expression exists for $y$), where $r_i$s are randomly chosen during initialisation (they are fixed for $x$ and $y$). Using a similar analysis as that in the Freivald's algorithm~\cite[Ch~1(Verifying matrix multiplication)]{upfal},
\begin{align*}
\Pr_{\rho,R}[\phi_j(x)=\phi_j(y) ~|~ F_j(x)\not=F_j(y)] = \tfrac{1}{p}. \numberthis \label{eq:prob_pi_r}   
\end{align*}

Due to Equations~\ref{eq:prob_pi}, \ref{eq:prob_pi_r}, we have
\begin{align*}
& \Pr_{\rho,R} [\phi_j(x)\not=\phi_j(y)] \\
= & \Pr_{\rho,R} [\phi_j(x)\not=\phi_j(y) ~|~ F_j(x)\not=F_j(y)] \cdot \Pr_{\rho,R} [F_j(x)\not=F_j(y)] \\
+ & \Pr_{\rho,R} [\phi_j(x)\not=\phi_j(y) ~|~ F_j(x)=F_j(y)] \cdot \Pr_{\rho,R} [F_j(x)=F_j(y)]\\
= & (1-\tfrac{1}{p})(1-(1-\tfrac{1}{d})^h).
\end{align*}
\end{proof}

The right-hand side of the expression in the statement of the lemma can be approximated as $(1-\tfrac{1}{p})\tfrac{h}{d}$ which is stated as case (a) of Theorem~\ref{thm:main-intro}.
The lemma also allows us to relate the Hamming distance of the sketches to the Hamming distance of the vectors which is our main tool to define an estimator.

\begin{restatable}{lem}{expectationlemma}\label{lem:expectation}
Let $h$ denote $HD(x,y)$ for two input vectors $x,y$ to \fsketch, $f$ denote $HD(\phi(x),\phi(y))$ and $f^*$ denote $\E[HD(\phi(x),\phi(y))]$. Then
\begin{align*}
f^* = \E[f] & = d\left( 1-\tfrac{1}{p}\right)\left(1 - \big(1-\tfrac{1}{d}\big)^h \right).
\end{align*}
\end{restatable}

The lemma is easily proved using Lemma~\ref{lem:expectation-previous} by applying the linearity of expectation on the number of coordinates $j$ such that $\phi_j(x)\not=\phi_j(y)$. We are now ready to define an estimator for the Hamming distance.

%
Using $D=\left(1 - \tfrac{1}{d} \right)$ and $P=\left(1-\tfrac{1}{p}\right)$, we
can write 
\[
f^*=dP(1-D^h) \mbox{ and } h=\ln\left( 1 - \tfrac{f^*}{dP} \right)/\ln D.
\numberthis\label{eqn:fstar}
\]

Our proposal to estimate $h$ is to obtain a tight approximation of $f^*$ and then use the above expression.

\begin{defi}[Estimator of Hamming distance]\label{defn:estimator}
Given sketches $\phi(x)$ and $\phi(y)$ of data points $x$ and $y$, suppose $f$ represents $HD(\phi(x),\phi(y))$.
We define the estimator of $HD(x,y)$ as
$\hat{h} = \ln\left( 1 - \tfrac{f}{dP} \right)/\ln D$ if $f < dP$ and $2\sigma$ otherwise.
\end{defi}

Observe that $\hat{h}$ is set to $2\sigma$ if $f \ge dP$. However, we shall show in the next section that this occurs very rarely.
%

\subsection{Analysis of Estimator}\label{subsec:analysis}

$\hat{h}$ is pretty reliable when the actual Hamming distance is 0; in that case 
$\phi(x)=\phi(y)$ and thus, $f=0$ and so is $\hat{h}$.
However, in general, $\hat{h}$ could be different from $h$. 
The main result of this section is that their difference can be upper bounded \bl{when we set} the dimension of \fsketch to $d=4\sigma$.

The results of this subsection rely on the following lemma that proves that an observed value of $f$ is concentrated around its expected value $f^*$.

\begin{restatable}{lem}{concentrationlemma}
\label{lem:concentration}Let $\alpha$ denote a desired additive accuracy. Then, for any $x,y$ with sparsity $\sigma$,\\
    \centerline{$\displaystyle \Pr\Big[|f - f^*| \ge \alpha\Big] \le 2\exp{(-\tfrac{\alpha^2}{4\sigma})}$.}
\end{restatable}

The proof of the lemma employs martingales and McDiarmid's inequality and is available in Appendix~\ref{appendix:subsec:analysis}. 
%
The lemma allows us to upper bound the probability of $f \ge dP$.

\begin{restatable}{lem}{upperboundf}\label{lem:upper-bound-f}
    $\Pr[f \ge dP] \le 2 \exp(-P^2\sigma)$.
\end{restatable}

The right-hand side is a very small number, e.g., it is of the order of $10^{-278}$ for $p=5$ and $\sigma=1000$. 
The proof is a straightforward application of Lemma~\ref{lem:concentration} and is explained in Appendix~\ref{appendix:subsec:analysis}. Now we are ready to show that the estimator $\hat{h}$, which uses $f$ instead of
$f^*$ (refer to Equation~\ref{eqn:fstar}) is almost equal to the actual Hamming distance.

\begin{restatable}{lem}{hconcentrationlemma}\label{lem:hconcentrationlemma}
    Choose $d=4\sigma$ as the dimension of \fsketch and choose a prime $p$ and
    an error parameter $\delta \in (0,1)$ (ensure that $1-\tfrac{1}{p} \ge
    \tfrac{4}{\sqrt{\sigma}} \sqrt{\ln \tfrac{2}{\delta}}$ --- see
    the proof for discussion). Then the estimator defined in
    Definition~\ref{defn:estimator} is close to the Hamming distance between $x$ and $y$ with high
    probability, i.e.,\\
    \centerline{$\displaystyle\Pr\big[|\hat{h}-h| \ge \tfrac{32}{1-1/p}\sqrt{\sigma \ln \tfrac{2}{\delta}} \big] \le \delta$.}
\end{restatable}

If the data vectors are not too dissimilar which is somewhat evident from Figure~\ref{fig:hamming_distance_distrib}, then a better compression is possible which is stated in the next lemma.
The proofs of both these lemmas are fairly algebraic and use standard inequalities; they are included in Appendix~\ref{appendix:subsec:analysis}.

\begin{restatable}{lem}{hconcentrationlemmatight}\label{lem:hconcentrationlemmatight}
    Suppose we know that $h \le \sqrt{\sigma}$ and choose $d=16\sqrt{\sigma \ln \tfrac{2}{\delta}}$
    as the dimension for \fsketch. Then (a) also $f < dP$ with high probability and moreover we get a better estimator. That is,
    (b) $\displaystyle\Pr\big[ |\hat{h} - h | \big] \ge \tfrac{8}{1-1/p}\sqrt{\sigma\ln
    \tfrac{2}{\delta}} \big] \le \delta $.
\end{restatable}





The last two results prove case (b) of Theorem~\ref{thm:main-intro} which states that the estimated Hamming distances are almost always fairly close to the actual Hamming distances. We want to emphasise that the above claims on $d$ and accuracy are only theoretical
bounds obtained by worst-case analysis. We show in our empirical evaluations
that an even smaller $d$ leads to better accuracy in practice for real-life instances.

There is a way to improve the accuracy even further by generating multiple \fsketch using several independently generated internal variables and combining the estimates obtained from each. We observed that the median of the estimates can serve as a good statistic, both theoretically and empirically. We discuss this in detail in Appendix~\ref{subsec:minfsketch}.

\subsection{Complexity analysis}\label{subsec:complexity}

\begin{table*}[]
    \centering
    \caption{Space savings offered by \fsketch on an example scenario with $2^{20}$ data points, each of $2^{10}$ dimensions but having only $2^7$ non-zero entries where non-zero entry belongs to one of $2^3$ categories. \fsketch dimension is $2^9$ (as prescribed theoretically) and its parameter $p$ is close to $2^5$. (*) The data required to construct the sketches is no longer required after the construction.\label{tab:space-example}}
    \begin{tabular}{cc||cc}
    \hline
    \multicolumn{2}{c||}{Uncompressed} & \multicolumn{2}{c}{Compressed}\\
    \hline
    Naive & Sparse vector format & \fsketch construction (*) & Storage of sketches \\
    \hline
	$2^{20} \times 2^{10} \times 3$     & $2^{20} \times 2^7 \times
	(\log 2^3 + \log 2^{10})$ & $2^{10} \times (\log 2^9 + \log 2^5) + 5$ & $2^{20} \times \log (2^5)$\\
	\hline
    \end{tabular}
\end{table*}

The results in the previous section show that the accuracy of the estimator $\hat{h}$ can be
tightened, or a smaller probability of error can be achieved, by choosing large values of $p$ which has a downside of a larger storage requirement. In this section, we discuss these dependencies and other factors that affect the complexity of our proposal.

The USP of \fsketch is its efficiency.
There are two major operations with respect to \fsketch --- construction of sketches and estimation of Hamming distance from two sketches. Their time and space requirements are given in the following table and explained in detail in Appendix~\ref{subsec:appendix-complexity}.

\noindent\begin{tabular}{lclc}
    \hline
    Construction & & Estimation & \\
    \hline
    time per sketch & $O(n)$ & time per pair & $O(d \log p)$\\
    space per sketch & $O(d \log p)$ & & \\
    \hline
\end{tabular}\\

\indent We are aware of efficient representations of sparse data vectors, but for the sake of simplicity we assume full-size arrays to store vectors in this table; similarly, we assume simple dictionaries for storing the internal variables $\rho,R$ and $p$. While it may be possible to reduce the number of random bits by employing $k$-wise independent bits and mappings, we left it out of the scope of this work.  


Both the operations are quite fast compared to the matrix-based and learning-based methods. There is very little space overhead too; we explain the space requirement with the help of an example in Table~\ref{tab:space-example} --- one should keep in mind that a sparse representation of a vector has to store the non-zero
entries as well as their positions in it.

Apart from the efficiency in both time and space measures, \fsketch provides additional benefits. \bl{Recall that each entry of an \fsketch is an integral value from 0 to $p-1$. Even though $0$ does not necessarily indicate a missing feature in a compressed vector, we show below that $0$ has a predominant presence in the sketches. The sketches can therefore be treated as sparse vectors that further facilitates their efficient storage.}

\begin{restatable}{lem}{sparsitylemma}\label{lem:sparsity}
    If $d=4\sigma$ (as required by Lemma~\ref{lem:hconcentrationlemma}), then the expected number of non-zero entries of $\phi(x)$ is upper bounded by $\tfrac{d}{4}$. Further, at least 50\% of $\phi(x)$ will be zero with probability at least $\tfrac{1}{2}$.
\end{restatable}
The lemma can be proved using a balls-and-bins type analysis (see Appendix~\ref{appendix:subsec:complexity} for the entire proof).

\subsection{Sketch updating}

Imagine a situation where the categories of attributes can change dynamically, and they can both ``increase'', ``decrease'' or even ``vanish''. We present Algorithm~\ref{alg:update} to incorporate such changes {\em without recomputing the sketch afresh.} The algorithm simply uses the formula for a sketch entry as given in Observation~\ref{obs:formula}.

Most hashing-based sketching and dimensionality reduction algorithms that we have encountered either require complete regeneration of $\phi(x)$ when some attributes of $x$ change or are able to handle addition of previously missing attributes but not their removal.
    \begin{algorithm}[t]
	\noindent\hspace*{\algorithmicindent} \textbf{input:} data
	vector $x$ and its existing sketch $\phi(x)=\langle \phi_1(x),
	\phi_2(x), \ldots \phi_d(x) \rangle$\\
	\noindent\hspace*{\algorithmicindent} \textbf{input:} change $x_i:v
	\mapsto v'$ \hfill $\rhd \quad v'$ can be any value in $\{0,1,\ldots, c\}$\\
	\noindent\hspace*{\algorithmicindent} \textbf{parameters:} $\rho,R=[r_1 \ldots r_n], p$
	(same as that was used for generating the sketch) 
	\begin{algorithmic}[1]
	    \State $j=\rho(i)$
	    \State update $\phi_(x) = \big( \phi_j(x) + (v'-v)\cdot r_i
	    \big) \mod{p}$
	    \State \Return updated $\phi(x)$
	\end{algorithmic}
	\caption{Update sketch $\sigma(x)$ of $x$ after $i$-th attribute of
	$x$ changes from $v$ to $v'$ \label{alg:update}}
    \end{algorithm}

%

\section{Experiments}\label{sec:experiments}

We performed our experiments on a machine having {Intel(R) Xeon(R) CPU E5-2650 v3 @ 2.30GHz, 94 GB RAM, and running a Ubuntu 64-bits OS.}


We first study the effect of the internal parameters of our proposed solution on its performance. We start with the effect of the prime number $p$; then we compare \fsketch with the appropriate baselines for several unsupervised data-analytic tasks (see Table~\ref{tab:baseline_characteristic}) and objectively establish these advantages of \fsketch over the others.

\begin{enumerate}[(a)]
    \item Significant speed-up in the dimensionality reduction time,
    \item considerable savings in the time for the end-tasks (e.g., clustering) which now runs on the low-dimensional sketches,
    \item but with comparable accuracy of the end-tasks (e.g., clustering).
\end{enumerate}

Several baselines threw {\it out-of-memory} errors or did not stop on certain datasets. We discuss the errors separately in Section~\ref{sec:appendix_oom_error} in Appendix.

\subsection{Dataset description}
The efficacy of our solution is best described for high-dimensional datasets. Publicly available categorical datasets being mostly low-dimensional, we treated several integer-valued freely available real-world datasets as categorical. Our empirical evaluation was done on the following seven such datasets with dimensions between $5000$ and $1.3$ million, and sparsity from $0.07\%$ to 30\%.
\begin{itemize}
    \item \texttt{Gisette Data Set}~\cite{UCI,Gisette}: 
	This dataset consists of integer feature vectors corresponding to images of handwritten digits and was constructed from the MNIST data. Each image, of $28 \times 28$ pixels, has been pre-processed (to retain the pixels necessary to disambiguate the digit $4$ from $9$) and then projected onto a higher-dimensional feature space represented to construct a 5000-dimension integer vector. 
   
    \item \texttt{BoW (Bag-of-words)}~\cite{UCI,DeliciousMIL}: We consider the following five corpus -- NIPS full papers, KOS blog entries, Enron Emails, NYTimes news articles, and tagged web pages from the social bookmarking site \texttt{delicious.com.} These datasets are ``BoW"(Bag-of-words) representations of the corresponding text corpora. In all these datasets, the attribute takes integer values which we consider as categories.
    \item \texttt{1.3 Million Brain Cell Dataset}~\cite{genomics20171}: This dataset contains the result of a single cell RNA-sequencing \texttt{(scRNA-seq)} of $1.3$ million cells captured and sequenced from an \texttt{E18.5} mouse brain~\footnote{\url{https://support.10xgenomics.com/single-cell-gene-expression/datasets/1.3.0/1M_neurons}}. Each gene represents a data point and for every gene, the dataset stores the read-count of that gene corresponding to each cell -- these read-counts form our features.
\end{itemize}

We chose the last dataset due to its very high dimension and the earlier ones due to their popularity in dimensionality-reduction experiments. \bl{We consider all the data points for KOS, Enron, Gisette, DeliciousMIL, a $10,000$ sized sample for NYTimes, and a $2000$ sized samples for BrainCell}. We summarise the dimensionality, the number of categories, and the sparsity of these datasets in the Table~\ref{tab:datasets}.

\begin{table}
\centering
  \caption{ Datasets}
  \label{tab:datasets}
  \resizebox{0.48\textwidth}{!}{%
  \noindent\begin{tabular}{lcccc}
  \hline
     Datasets & Categories & Dimension & Sparsity & \bl{No. of points} \\
  \hline
    \textrm{Gisette}~\cite{Gisette,UCI} & $999$ & $5000$ & $1480$&$\bl{13500}$\\
    \textrm{Enron Emails}~\cite{UCI} & $150$ & $28102$ & $2021$&$\bl{39861}$\\
    \textrm{DeliciousMIL}~\cite{DeliciousMIL,UCI} & $58$ & $8519$ & $200$&$\bl{12234}$\\
    \raggedright\textrm{NYTimes articles}~\cite{UCI} & $114$ & $102660$ & $ 871$&$\bl{10000}$\\
    \bl{\textrm{NIPS full papers}~\cite{UCI} }& \bl{$132$} & \bl{$12419$} & \bl{$914$} & \bl{$1500$}\\
    \raggedright\textrm {KOS blog entries}~\cite{UCI} & $42$ & $6906$ & $457$&$\bl{3430}$\\
    \textrm {Million Brain Cells from E18 Mice}~\cite{genomics20171} & $2036$ & $1306127$ & $1051$&$\bl{2000}$\\
  \hline
\end{tabular}%
}
\end{table}
 

\begin{table}[t]
\centering
  \caption{ 13 baselines}
  \label{tab:baselines_methods}
\begin{tabular}[t]{rrl}
\hline
1.&  {\rm SSD} &\textrm{Sketching via Stable Distribution}~\cite{SSD}\\
2.&    {\rm OHE} & \textrm{One Hot Encoding+BinSketch~\cite{ICDM}}\\
3.&    {\rm FH} & \textrm{Feature Hashing}~\cite{WeinbergerDLSA09}\\
4.&    {\rm SH} & \textrm{Signed-random projection/SimHash}~\cite{simhash}\\
5.&      {\rm KT} & \textrm{Kendall rank correlation coefficient}~\cite{kendall1938measure}\\
6.&    {\rm LSA} & \textrm{Latent Semantic Analysis}~\cite{LSI}\\
7.&    {\rm LDA} & \textrm{Latent Dirichlet Allocation}~\cite{LDA}\\
8.&    {\rm MCA} & \textrm{Multiple Correspondence Analysis}~\cite{MCA}\\
9.&    {\rm NNMF} & \textrm{Non-neg. Matrix Factorization}~\cite{NNMF}\\
10.&    {\rm PCA} & Vanilla \textrm{Principal component analysis}\\
11.&    \bl{ {\rm VAE}} &  \bl{\textrm{Variational autoencoder}}~\cite{Kingma2014}\\
12.&    \bl{ {\rm CATPCA}} &  \bl{\textrm{Categorical PCA}}~\cite{Sulc2015DimensionalityRO}\\
13.&    \bl{ {\rm HCA}} &  \bl{\textrm{Hierarchical Cluster Analysis}}~\cite{Sulc2015DimensionalityRO}\\
    
\hline
\end{tabular}
\end{table}

\begin{table*}[!t]
\centering
\caption{{ Summarisation of the baselines. } }\label{tab:baseline_characteristic}
\resizebox{\textwidth}{!}{%
  \begin{threeparttable}
  \begin{tabular}{Slcccccccccccccc}
    \toprule
    Characteristics&FSketch&FH&SH&SSD&OHE&KT&NNMF&MCA&LDA&LSA&PCA&{VAE}&{CATPCA}&{HCA}\\
    \midrule
\pbox{5em}{Output\\ discrete\\ sketch}&\cmark&\cmark&\cmark&\xmark&\cmark&\cmark&\xmark&\xmark&\xmark&\xmark&\xmark&\xmark &\xmark&\cmark\\[2em]
\pbox{5em}{Output\\ real-valued\\ sketch}&\xmark&\xmark&\xmark&\cmark&\xmark&\xmark&\cmark&\cmark&\cmark&\cmark&\cmark&\cmark&\cmark&\xmark\\[2em]
\pbox{5em}{Approximating\\ distance\\ measure}&{Hamming}&\pbox{4em}{ Dot\\product}& {Cosine}& {Hamming}&{Hamming}& \texttt{NA} &\texttt{NA} &\texttt{NA} &\texttt{NA} & \texttt{NA}& \texttt{NA} & \texttt{NA} &\texttt{NA}&\texttt{NA}\\[2em]
\pbox{5em}{Require \\labelled data}&\xmark&\xmark&\xmark&\xmark&\xmark&\xmark&\xmark&\xmark&\xmark&\xmark&\xmark&\xmark&\xmark&\xmark\\[2em]
\pbox{5em}{Dependency\\ on the size\\ of sample \tnote{*}} &\xmark&\xmark&\xmark&\xmark&\xmark&\xmark&\xmark &\cmark&\xmark&\cmark &\cmark&\xmark&\cmark&\xmark\\[2em]
\pbox{5em}{End tasks\\ comparison}&All&All&All&All&All&All& \pbox{5em}{Clustering,\\ Similarity Search}
&\pbox{5em}{Clustering,\\ Similarity Search}
&\pbox{5em}{Clustering,\\ Similarity Search}
&\pbox{5em}{Clustering,\\ Similarity Search}
&\pbox{5em}{Clustering,\\ Similarity Search}
&\pbox{5em}{Clustering,\\ Similarity Search}
&\pbox{5em}{Clustering,\\ Similarity Search}&\pbox{5em}{Clustering,\\ Similarity Search}\\
\bottomrule
 \end{tabular}
 \begin{tablenotes}
 \item[*] The size of the maximum possible reduced dimension is the minimum of the number of data points and the dimension.
 \end{tablenotes}
 \end{threeparttable}
}
\end{table*}

\subsection{Baselines}
Recall that \fsketch (hence \minfsketch) {\em compresses
categorical vectors to shorter categorical vectors in an unsupervised manner that	``preserves'' Hamming distances}. 

Our first baseline is based on one-hot-encoding (OHE) which is one of the most common methods to convert categorical data to a numeric vector and can approximate pairwise Hamming distance (refer to Appendix~\ref{appendix:OHE+BS}). Since OHE actually increases the dimension to very high levels (e.g., the dimension of the binary vectors obtained by encoding the NYTimes dataset is $11,703,240$), the best way to use it is by further compressing the one-hot encoded vectors. For empirical evaluation we applied BinSketch~\cite{ICDM} which is the state-of-the-art binary-to-binary dimensionality reduction technique that preserves Hamming distance. We refer to the entire process of OHE followed by BinSketch  simply by OHE in the rest of this section.


To the best of our knowledge, there is no sketching algorithm other than OHE that compresses high-dimensional categorical vectors to low-dimensional categorical (or integer) vectors that preserves the original pairwise Hamming distances. Hence, we chose as baseline state-of-the-art and popularly employed algorithms that either preserve Hamming distance or output discrete-valued sketches (preserving some other similarity measure).  
We list them in Table~\ref{tab:baselines_methods} and  tabulate their characteristic in Table~\ref{tab:baseline_characteristic}. Their implementation details are discussed in Appendix~\ref{sec:Reproducibility_baseline}.


\begin{figure*}
\centering
\includegraphics[width=\linewidth]{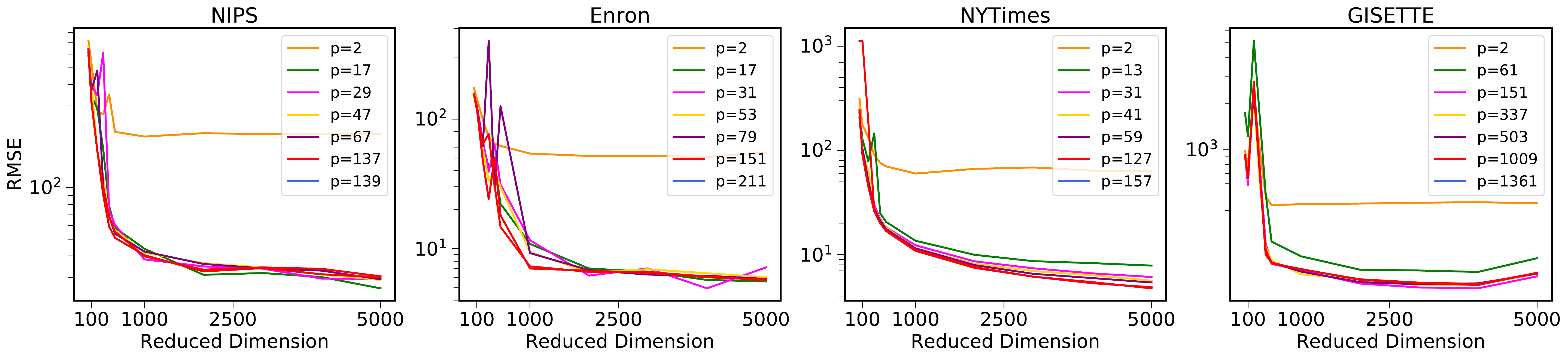}
\caption{{Comparison of $\RMSE$ measure obtained from \texttt{FSketch} algorithm on various choices of  $p$. \bl{Values of $c$ for NIPS, Enron, NYTimes, and GISETTE are 132, 150, 114, and 999, respectively.}
}}
\label{fig:varying_p}
\end{figure*}

 We include Kendall rank correlation coefficient (KT)~\cite{kendall1938measure} -- a feature selection algorithm which generates discrete valued sketches. Note that if we apply Feature Hashing (FH), SimHash (SH), and KT naively on  categorical datasets, we get discrete valued sketches on which Hamming distance can be computed.
We also include a few other well known dimensionality reduction methods such as Principal component analysis (PCA), Non-negative Matrix Factorisation (NNMF)~\cite{NNMF}, Latent Dirichlet Allocation (LDA)~\cite{LDA}, Latent Semantic Analysis (LSA)~\cite{LSI}, \bl{Variational Autoencoder (VAE) \cite{Kingma2014}, Categorical PCA (CATPCA)\cite{Sulc2015DimensionalityRO}, Hierarchical Cluster Analysis (HCA) \cite{Sulc2015DimensionalityRO}}   all of which output real-valued sketches.

\begin{figure*}
\centering
\includegraphics[width=\linewidth]{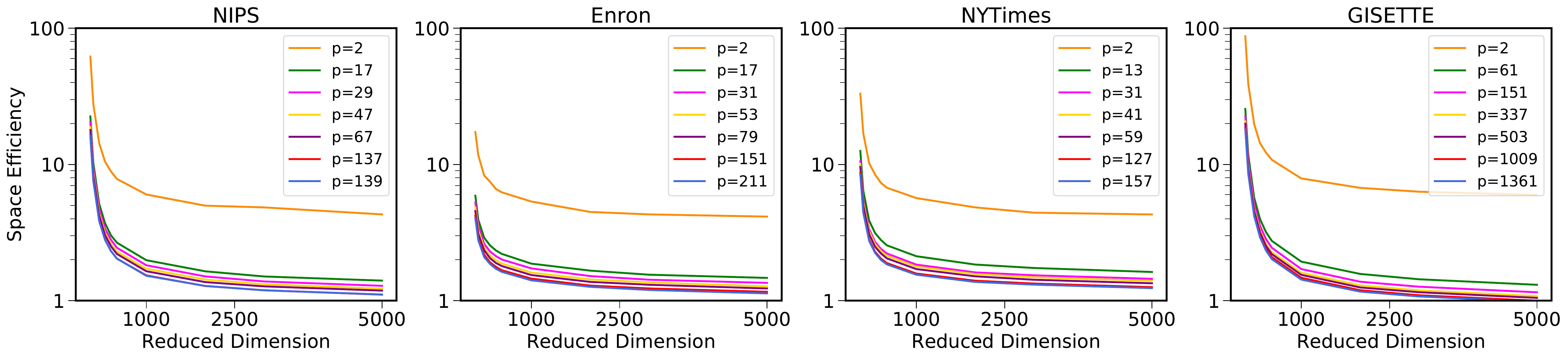}
    \caption{{Space overhead of uncompressed vectors stored as a list of non-zero entries and their positions. $Y$-axis represents the ratio of the space used by uncompressed vector to that obtained from \fsketch.
}}
\label{fig:varying_p_space}
\end{figure*}

\begin{figure*}
\centering
\includegraphics[width=\linewidth]{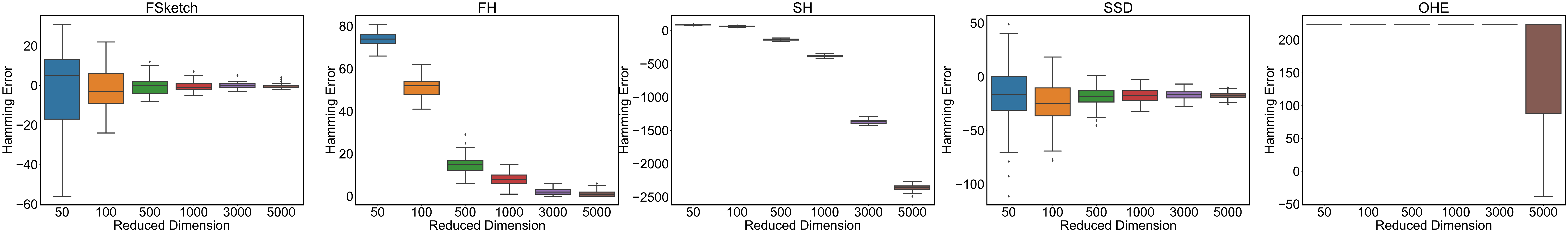}
    \caption{{Comparison of avg.\ error in estimating Hamming distance of a pair of points from the Enron dataset.}}

\label{fig:box_plot_hamming_error}
\end{figure*}

\subsection{Choice of $p$}
We discussed in Section~\ref{sec:analysis} that a larger value of $p$ (a prime number) leads to a tighter estimation of Hamming distance but degrades sketch sparsity, which negatively affects performance at multiple fronts, and moreover, demands more space to store a sketch. We conducted an experiment to study this trade-off, where we ran our proposal with different values of $p$, and computed the corresponding $\RMSE$ values. The $\RMSE$ is defined as the square-root of the average error, among all pairs of data points, between their actual Hamming distances and the corresponding estimate obtained via \texttt{FSketch}. Note that a lower $\RMSE$ indicates that the sketch correctly estimates the underlying pairwise Hamming distance.  We also note the corresponding \textit{space overhead} which is defined as the ratio of the space used by uncompressed vector and its sketch obtained from~\texttt{FSketch}. We consider storing a data point in a typical sparse vector format -- a list of non-zero entries and their positions (see Table~\ref{tab:space-example}). We summarise our results in Figures~\ref{fig:varying_p} and  ~\ref{fig:varying_p_space}, respectively. We observe that  a large value of $p$ leads to a lower $\RMSE$ (in Figure~\ref{fig:varying_p}), however simultaneously it leads to a smaller space compression (Figure~\ref{fig:varying_p_space}). \bl{
As a heuristic, we decided to set $p$ as the next prime after $c$ as shown in this table.}\\
{
 \footnotesize
\noindent\begin{tabular}{lc@{\hskip 2em}lc@{\hskip 2em}lc}
    \toprule
     Brain cell & $2039$ & NYTimes & $127$ & Enron & $151$ \\
     KOS & $43$ & Delicious & $59$ & Gisette & $1009$\\
     \bl{NIPS} &\bl{$137$}&&&\\ 
     \bottomrule
\end{tabular}
}

\bl{That said, the experiments reveal that, at least for the datasets in the above experiments, setting $p$ to be at least $c/4$ may be practically sufficient, since there does not appear to be much advantage in using a larger $p$.}

\begin{figure*}
{
\centering
\includegraphics[width=\linewidth]{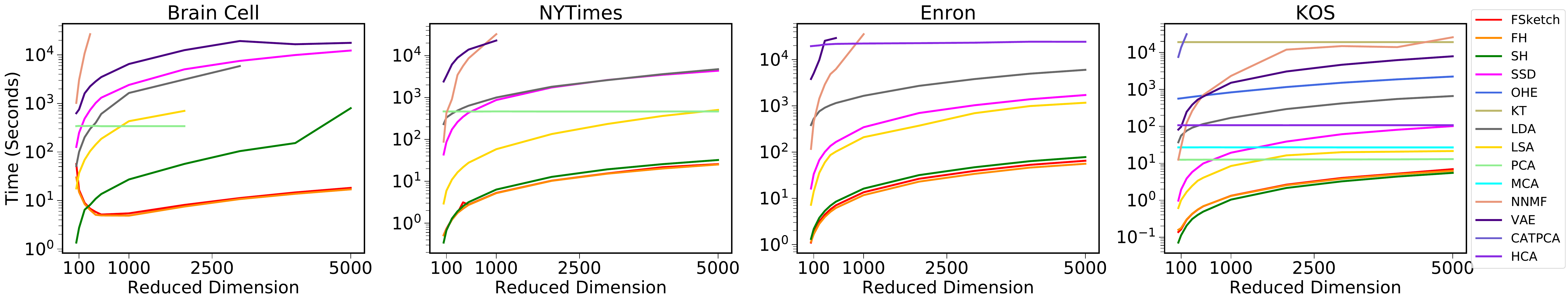}
    \caption{Comparison among the baselines on the dimensionality reduction time. See Appendix~\ref{appendix:section:extended_exp} for results on the other datasets which show a similar trend and Section~\ref{sec:appendix_oom_error} for the errors encountered by some baselines.}
\label{fig:reduction_time}
}
\end{figure*}


\begin{table*}
{
\centering
    \caption{{Speedup of \texttt{FSketch} \textit{w.r.t.} baselines on the reduced dimension $1000$. 
    \texttt{OOM} indicates ``out-of-memory error'' and {\tt DNS} indicates ``did not stop'' after a sufficiently long time.
} }\label{tab:speed_up_dim_time}
\scalebox{0.99}{%
  \begin{tabular}{ lccccccccccccr}
    \toprule 
    Dataset  &   OHE          &       KT       &      NNMF       &    MCA        &     LDA     &     LSA       &   PCA         &   VAE         &     SSD            &     SH       &    FH        & CATPCA & HCA   \\
    \midrule
NYTimes      &$ \texttt{OOM} $&$ \texttt{OOM}  $&$ 6149\times $&$  \texttt{OOM} $&$ 189\times $&$  11.5\times $&$ 88.14\times $&$ 4340 \times $&$ 164.9\times $&$ 1.2\times   $&$ 0.99\times$     &${\tt DNS}$&${\tt DNS}$ \\
Enron        &$ \texttt{OOM} $&$ \texttt{DNS}  $&$2624\times$ &$  \texttt{OOM} $&$ 122\times $&$ 15.5 \times $&$\texttt{OOM} $&$ \texttt{DNS}$&$ 25.5\times   $&$ 1.25\times  $&$ 0.87\times$     &${\tt DNS}$&$1268.2\times$\\ 
KOS          &$ 629\times    $&$ 14455\times   $&$ 1754\times    $&$   20.41\times $&$ 128\times $&$ 6.40\times  $&$ 9.5\times $&$  1145\times $&$ 14.62\times   $&$ 0.79\times  $&$ 0.98\times$  &${\tt DNS}$&$81.24\times$\\
DeliciousMIL &$ 1332\times   $&$ 14036\times   $&$ 1753\times    $&$   40.39\times $&$ 136\times $&$ 6.6\times   $&$ 18.1\times $&$  1557\times $&$ 29.2 \times $  &$ 0.61\times  $&$ 0.90\times$ &${\tt DNS}$&$117.6\times$\\
Gisette      &$ 399\times    $&$ 1347\times    $&$ 459\times  $&$   5.7\times   $&$ 269\times $&$ 5.4\times   $&$ 4.2\times   $&$  285\times  $&$  8.1\times    $&$ 0.69\times  $&$ 0.98\times$   &${\tt DNS}$&$16.78\times$\\ 
NIPS         &$ 378\times    $&$ 15863\times    $&$ 1599\times  $&$  26.6\times   $&$ 302\times $&$ 6.4\times   $&$ 3.17\times   $&$  451\times  $&$  29.9\times    $&$ 0.47\times  $&$1.20\times$ &${\tt DNS}$&$58.49\times$\\ 
Brain Cell     &$ \texttt{OOM} $&$\texttt{OOM} $&$ {\tt DNS}  $&$ \texttt{OOM} $&$ 322\times $&$ 79.38\times  $&$ 62.7\times $&$1198\times$    &$ 443 \times$ &  $5\times$ & $0.89\times$  &${\tt DNS}$&${\tt DNS}$\\
\bottomrule
 \end{tabular}
 }
}
\end{table*}

\begin{figure*}
\centering
\includegraphics[width=\linewidth]{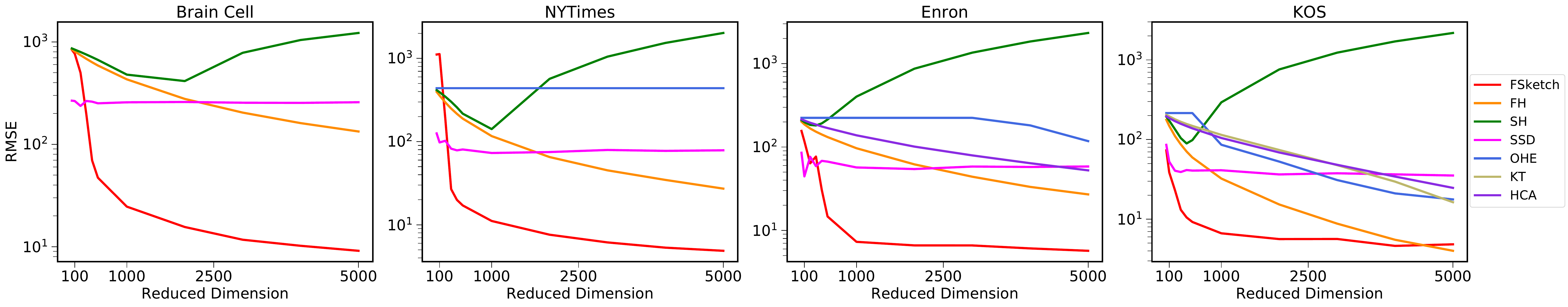}
\caption{{Comparison on $\RMSE$ among baselines. A lower value is an indication of better performance.  See Appendix~\ref{appendix:section:extended_exp} for results on the other datasets which show a similar trend. 
}}
\label{fig:rmse}
\end{figure*}

\subsection{Variance of \fsketch}
In Section~\ref{subsec:analysis} we explained that the bias of our estimator is upper bounded with a high likelihood. However, there remains the question of its variance. To decide the worthiness of our method, we compared the variance of the estimates of the Hamming distance obtained from \fsketch and from the other randomised sketching algorithms with integer-valued sketches (KT was not included as it is a deterministic algorithm, and hence, has zero variance).

Figure~\ref{fig:box_plot_hamming_error} shows the Hamming error (estimation error) for a randomly chosen pair of points from the Enron dataset, averaged over 100 iterations. 
We make two observations.

First is that the estimate using \texttt{FSketch} is closer to the actual Hamming distance even at a smaller
reduced dimension; in fact, as the reduced dimension is increased, the variance
becomes smaller and the Hamming error converges to zero. Secondly, \fsketch causes a smaller error compared to the other baselines. On the other hand, {feature hashing} highly underestimates the actual Hamming distance, but has low variance, and tends to have negligible Hamming error with an increase of the reduced dimension. The behaviour of {SimHash} is counter-intuitive as on lower reduced dimensions it closely estimates the actual Hamming distances, but on larger dimensions it starts to highly underestimate the actual Hamming distances. This creates an ambiguity on the choice of a dimension for generating a low-dimensional sketch of a dataset.
 Similar to \fsketch, the sketches produced by SSD, though real-valued, allow estimation of pairwise Hamming distances. However the estimation error increases with the reduced dimension. Lastly, OHE seems to be highly underestimating pairwise Hamming distances.

\subsection{Speedup in dimensionality reduction}\label{subsubsec:dim_red_time}
 We compress the datasets to several dimensions using \fsketch and the baselines and report their running times in Figure~\ref{fig:reduction_time}. We notice that \fsketch has a comparable speed \textit{w.r.t.} Feature hashing and SimHash, and is significantly faster than the other baselines. However, both feature hashing and SimHash are not able to accurately estimate the Hamming distance between data points and hence perform poorly on $\RMSE$ measure (Subsection~\ref{subsubsec:rmse}) and the other tasks. Many baselines such as OHE, KT, NNMF, MCA, \bl{CATPCA, HCA} give \textit{``out-of-memory”} (OOM) error, and also didn’t stop (DNS) even after running for a sufficiently long time ($\sim~10$ hrs) on high dimensional datasets such as Brain Cell and NYTimes. On other moderate dimensional datasets such as Enron and KOS, our speedup \textit{w.r.t.} these baselines are of the order of a few thousand. We report the numerical speedups that we observed in Table~\ref{tab:speed_up_dim_time}.

\begin{figure*}
{
\centering
    \includegraphics[width=\linewidth]{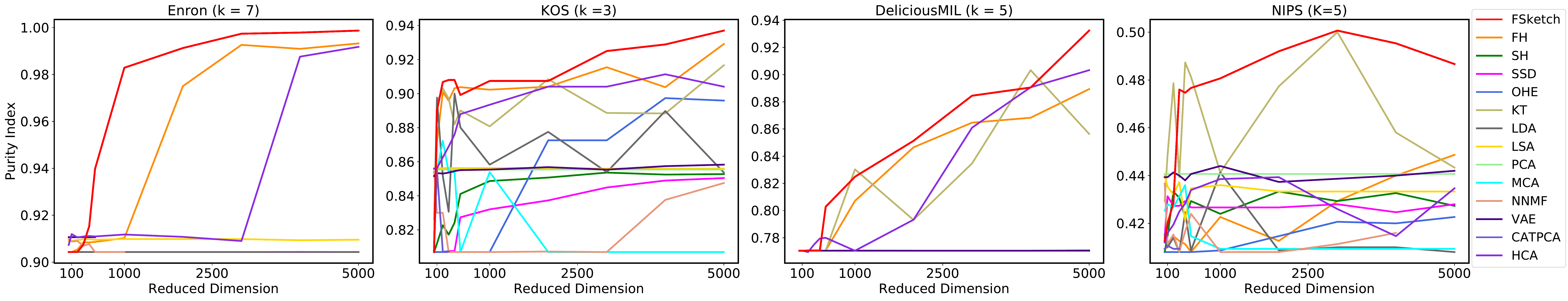}
\caption{{Comparing the quality of clusters on the compressed datasets.  See Appendix~\ref{appendix:section:extended_exp} for results on the other datasets which show a similar trend.
}}
\label{fig:purity}
}
\end{figure*}
 
 \begin{figure*}
 {
\centering
\includegraphics[width=\linewidth]{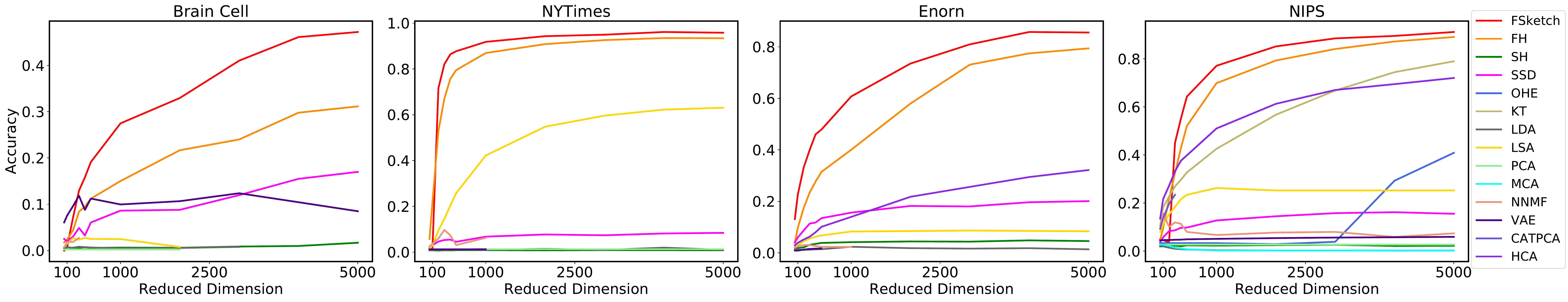}
\caption{{Comparing the performance of the similarity search task (estimating top-$k$ similar points with $k=100$) achieved on the reduced dimensional data obtained from various baselines.  See Appendix~\ref{appendix:section:extended_exp} for results on the other datasets which show a similar trend.
}}
\label{fig:similarity_search}
}
\end{figure*}

\subsection{Performance on root-mean-squared-error (RMSE)}\label{subsubsec:rmse}
How good are the sketches for estimating Hamming distances between the uncompressed points in practice? To answer this, we compare \fsketch with integer-valued sketching algorithms, namely,  {feature hashing}, {SimHash}, {Kendall correlation coefficient} and OHE+BinSketch.   Note that {feature hashing} and {SimHash} are known to approximate inner product and cosine similarity, respectively. However, we consider them in our comparison nonetheless as they output discrete sketches and Hamming distance can be computed on their sketch. We also include SSD for comparison which outputs real-valued sketches and  estimates original pairwise Hamming distance. 
For each of the methods we compute its $\RMSE$ as the square-root of the average error, among all pairs of data points, between their actual Hamming distances and their corresponding estimates (for \fsketch the estimate was obtained using Definition~\ref{defn:estimator}). Figure~\ref{fig:rmse} compares these values of $\RMSE$ for different dimensions; note that a lower $\RMSE$ is an indication of better performance. It is immediately clear that the $\RMSE$ of \texttt{FSketch} is the lowest among all; furthermore, it falls to zero rapidly with increase in reduced  dimension.  This demonstrates that our proposal \texttt{FSketch} estimates the underlying  pairwise Hamming distance better than the others.

\begin{table}[t]
{
\caption{{
Speedup from running tasks on 1000-dimensional sketches instead of the full
    dimensional dataset. We got a DNS error while running clustering on the uncompressed BrainCell dataset. 
    } }\label{tab:speed_up_task_time}
 \centering
\resizebox{0.48\textwidth}{!}{%
\addtolength\tabcolsep{-4pt}
  \begin{tabular}{lccccccc}
    \toprule
    Task           & Brain cell      & NYTimes        &   Enron            &  NIPS         &  KOS          &   Gisette     &    DeliciousMIL\\
    \midrule
Clustering         &$NA   $&$ 139.64\times  $&$  21.15\times    $&$ 10.6\times  $&$ 3.93\times  $&$ 4.35\times  $&$  5.84\times  $\\
Similarity Search  &$    1231.6\times  $&$  118.12\times $&$ 48.15\times  $&$ 15.1\times  $&$ 10.56\times  $&$ 8.34\times  $&$ 17.76\times $\\
 \bottomrule
  \end{tabular}
  \addtolength\tabcolsep{4pt}
  }
  
  }
\end{table}

\subsection{Performance on clustering}
We compare the performance of \fsketch with baselines on the task of clustering and similarity search, and present the results for the first task in this section. The objective of the clustering experiment 
was to test if the data points in the reduced dimension maintain the original clustering structure. If they do, then it will be immensely helpful for those techniques that use a clustering, e.g., spam filtering. We used the {\it purity index} to measure the quality of $k$-mode and $k$-means clusters on the reduced datasets obtained through the compression algorithms; the ground truth was obtained using $k$-mode on the uncompressed data (for more details refer to Appendix~\ref{appendix:clustering}).



We summarise our findings on quality in Figure~\ref{fig:purity}. \bl{ The compressed versions of the NIPS, Enron, and KOS  datasets that were obtained from \fsketch yielded the best purity index as compared to those obtained from the other baselines; for the other datasets the compressed versions from \fsketch are among the top. Even though it appears that KT offers comparable performance on the KOS, DeliciousMIL, and Gisette datasets \textit{w.r.t.}  \texttt{FSketch}, the downside of using KT is that its 
compression time is much higher than that of \texttt{FSketch} (see Table~\ref{tab:speed_up_dim_time}) on those datasets, and moreover it gives OOM/DNS error on the remaining datasets. Performance of FH also remains in the top few. However, its performance degrades on the NIPS dataset.}

We tabulate the speedup of clustering of \fsketch-compressed data over uncompressed data in Table~\ref{tab:speed_up_task_time} where we observe significant speedup in the clustering time, e.g.,  \bl{$139\times$} when run on a $1000$ dimensional \fsketch.

Recall that the dimensionality reduction time of our proposal is among the fastest among all the baselines which further reduces the total time to perform clustering by speeding up the dimensionality reduction phase. 
Thus the overall observation is that \fsketch appears to be the most suitable method for clustering among the current alternatives, especially, for high-dimensional datasets on which clustering would take a long time.

\subsection{Performance on similarity search}
We take up another unsupervised task -- that of similarity search. The objective here is to show that after dimensionality reduction the similarities of points with respect to some query points are maintained. To do so, we randomly split the dataset in two parts $5\%$ and $95\%$  -- the smaller partition is referred to as the \textit{query partition} and each point of this partition is called a query vector; we call the larger partition as \textit{training partition}. For each query vector, we find top-$k$ similar points in the training partition. We then perform dimensionality reduction using all the methods (for various values of reduced dimensions). Next, we process the compressed dataset where, for each query point, we compute the  top-$k$ similar points in the corresponding low-dimensional version of the training points, by maintaining the same split. For each query point, we compute the accuracy of baselines by taking the Jaccard ratio between the set of top-$k$ similar points obtained in full dimensional data with  the top-$k$ similar points obtained in reduced dimensional dataset. We repeat this for all the points in the querying partition, compute the average, and report this as accuracy.

We summarise our findings in Figure~\ref{fig:similarity_search}. Note that PCA, MCA and LSA can reduce the data dimension up to the minimum of the number of data points and the original data dimension. Therefore their reduced dimension is at most $2000$ \bl{ for Brain cell dataset}.

The top few methods appear to be feature hashing (FH), Kendall-Tau (KT), \bl{HCA} along with \fsketch. However, KT give \textit{OOM} and \textit{DNS} on the Brain cell, NYTimes and Enron datasets, and \bl{HCA give DNS error on BrainCell and NYTimes datasets}.  Ffurther, their dimensionality reduction time are much worse than \texttt{FSketch} (see Table~\ref{tab:speed_up_dim_time}). 


\bl{\fsketch outperforms FH on the BrainCell and the Enron datasets; however, on the remaining datasets, both of them 
 appear neck to neck for similarity search despite the fact that there is no known theoretical understanding of FH for Hamming distance --- in fact, it was included in the baselines as a heuristic because it offers discrete-valued sketches on which Hamming distance can be calculated. 
 Here want to point out that FH was not a consistent top-performer for clustering and similarity search.}

The two other methods that are designed for Hamming distance, namely SSD and OHE, perform significantly  worse than FSketch; in fact, the accuracy of OHE lies almost to the bottom on all the four datasets.

\bl{We also summarise the speedup of \fsketch-compressed data over uncompressed data, on similarity search task,  in Table~\ref{tab:speed_up_task_time}. We observe a significant speedup --  \textit{e.g.} \bl{$1231.6\times$} speedup on the BrainCell dataset when run on a   $1000$ dimensional \fsketch. }

To summarise, \fsketch is one of the best approaches towards similarity search for high-dimensional datasets and the best if we also require theoretical guarantees or applicability towards other data analytic tasks.

\section{Conclusion}\label{sec:conclusion}
In this paper, we proposed a sketching algorithm named \fsketch for sparse categorical data such that the Hamming distances estimated from the sketches closely approximate the original pairwise Hamming distances. The low-dimensional data obtained by \texttt{FSketch} are discrete-valued, and therefore, enjoy the flexibility of running the data analytics tasks suitable for categorical data. The sketches allow tasks like clustering, similarity search to run which might not be possible on a high-dimensional dataset.


Our method does not require learning from the dataset and instead, exploits randomization to bring forth large speedup and high-quality output for standard data analytic tasks. We empirically validated the performance of our algorithm on several metric and end tasks such as RMSE, clustering,  similarity search, and observed comparable performance while simultaneously getting significant speed up in dimensionality reduction and end-task with respect to several baselines.
\bl{A common practice to analyse high-dimensional datasets is to partition} them into smaller datasets. Given the simplicity, efficiency, and effectiveness of our proposal, we hope that \fsketch will allow \bl{such analysis to be done on the full datasets and on general-purpose hardware}.



\bibliographystyle{IEEEtran}


\begin{thebibliography}{10}
\providecommand{\url}[1]{#1}
\csname url@samestyle\endcsname
\providecommand{\newblock}{\relax}
\providecommand{\bibinfo}[2]{#2}
\providecommand{\BIBentrySTDinterwordspacing}{\spaceskip=0pt\relax}
\providecommand{\BIBentryALTinterwordstretchfactor}{4}
\providecommand{\BIBentryALTinterwordspacing}{\spaceskip=\fontdimen2\font plus
\BIBentryALTinterwordstretchfactor\fontdimen3\font minus
  \fontdimen4\font\relax}
\providecommand{\BIBforeignlanguage}[2]{{%
\expandafter\ifx\csname l@#1\endcsname\relax
\typeout{** WARNING: IEEEtran.bst: No hyphenation pattern has been}%
\typeout{** loaded for the language `#1'. Using the pattern for}%
\typeout{** the default language instead.}%
\else
\language=\csname l@#1\endcsname
\fi
#2}}
\providecommand{\BIBdecl}{\relax}
\BIBdecl

\bibitem{DNA}
J.~Moody, D.~T. (eds, M.~Kaufmann, M.~O. Noordewier, G.~G. Towell, and J.~W.
  Shavlik, ``Training knowledge-based neural networks to recognize genes in dna
  sequences,'' 1991.

\bibitem{HIV}
T.~Rognvaldsson, L.~You, and D.~Garwicz, ``State of the art prediction of hiv-1
  protease cleavage sites,'' \emph{Bioinformatics (Oxford, England)}, vol.~31,
  2014.

\bibitem{transaction}
W.~{Hämäläinen} and M.~{Nykänen}, ``Efficient discovery of statistically
  significant association rules,'' in \emph{2008 Eighth IEEE International
  Conference on Data Mining}, 2008, pp. 203--212.

\bibitem{itemset_categorical}
J.~Lavergne, R.~Benton, and V.~V. Raghavan, ``Min-max itemset trees for dense
  and categorical datasets,'' in \emph{Foundations of Intelligent Systems},
  L.~Chen, A.~Felfernig, J.~Liu, and Z.~W. Ra{\'{s}}, Eds.\hskip 1em plus 0.5em
  minus 0.4em\relax Berlin, Heidelberg: Springer Berlin Heidelberg, 2012, pp.
  51--60.

\bibitem{itemset_mining}
R.~Agrawal, T.~Imielinski, and A.~Swami, ``Mining association rules between
  sets of items in large databases,'' in \emph{SIGMOD '93: Proceedings of the
  1993 ACM SIGMOD international conference on Management of data}, 1993, pp.
  207--216.

\bibitem{web_transaction}
I.~Cadez, D.~Heckerman, C.~Meek, P.~Smyth, and S.~White, ``Visualization of
  navigation patterns on a web site using model-based clustering,'' in
  \emph{Proceedings of the Sixth ACM SIGKDD International Conference on
  Knowledge Discovery and Data Mining}, 2000, pp. 280--284.

\bibitem{image_categorical}
L.~Kurgan, K.~Cios, R.~Tadeusiewicz, M.~Ogiela, and L.~Goodenday, ``Knowledge
  discovery approach to automated cardiac spect diagnosis,'' \emph{Artificial
  intelligence in medicine}, vol.~23, pp. 149--69, 2001.

\bibitem{click_stream}
S.~Sidana, C.~Laclau, and M.-R. Amini, ``Learning to recommend diverse items
  over implicit feedback on pandor,'' 2018, pp. 427--431.

\bibitem{Hancock2020SurveyOC}
J.~T. Hancock and T.~M. Khoshgoftaar, ``Survey on categorical data for neural
  networks,'' \emph{Journal of Big Data}, vol.~7, pp. 1--41, 2020.

\bibitem{baraniuk2010low}
R.~G. Baraniuk, V.~Cevher, and M.~B. Wakin, ``Low-dimensional models for
  dimensionality reduction and signal recovery: A geometric perspective,''
  \emph{Proceedings of the IEEE}, vol.~98, no.~6, pp. 959--971, 2010.

\bibitem{plos_categorical_tips}
L.~H. Nguyen and S.~Holmes, ``Ten quick tips for effective dimensionality
  reduction,'' \emph{PLOS Computational Biology}, vol.~15, no.~6, pp. 1--19,
  2019.

\bibitem{chi_square}
\BIBentryALTinterwordspacing
H.~Liu and R.~Setiono, ``Chi2: feature selection and discretization of numeric
  attributes,'' in \emph{Seventh International Conference on Tools with
  Artificial Intelligence, {ICTAI} '95, Herndon, VA, USA, November 5-8, 1995},
  1995, pp. 388--391. [Online]. Available:
  \url{https://doi.org/10.1109/TAI.1995.479783}
\BIBentrySTDinterwordspacing

\bibitem{MI}
\BIBentryALTinterwordspacing
H.~Peng, F.~Long, and C.~H.~Q. Ding, ``Feature selection based on mutual
  information: Criteria of max-dependency, max-relevance, and min-redundancy,''
  \emph{{IEEE} Trans. Pattern Anal. Mach. Intell.}, vol.~27, no.~8, pp.
  1226--1238, 2005. [Online]. Available:
  \url{https://doi.org/10.1109/TPAMI.2005.159}
\BIBentrySTDinterwordspacing

\bibitem{kendall1938measure}
M.~G. Kendall, ``A new measure of rank correlation,'' \emph{Biometrika},
  vol.~30, no. 1/2, pp. 81--93, 1938.

\bibitem{scikit-learn}
F.~Pedregosa, G.~Varoquaux, A.~Gramfort, V.~Michel, B.~Thirion, O.~Grisel,
  M.~Blondel, P.~Prettenhofer, R.~Weiss, V.~Dubourg, J.~Vanderplas, A.~Passos,
  D.~Cournapeau, M.~Brucher, M.~Perrot, and E.~Duchesnay, ``Scikit-learn:
  Machine learning in {P}ython,'' \emph{Journal of Machine Learning Research},
  vol.~12, pp. 2825--2830, 2011.

\bibitem{kaggle_one_hot}
\BIBentryALTinterwordspacing
D.~Becker, ``Using categorical data with one hot encoding,'' 2018. [Online].
  Available:
  \url{https://www.kaggle.com/dansbecker/using-categorical-data-with-one-hot-encoding}
\BIBentrySTDinterwordspacing

\bibitem{ICDM}
\BIBentryALTinterwordspacing
R.~Pratap, D.~Bera, and K.~Revanuru, ``Efficient sketching algorithm for sparse
  binary data,'' in \emph{2019 {IEEE} International Conference on Data Mining,
  {ICDM} 2019, Beijing, China, November 8-11, 2019}, 2019, pp. 508--517.
  [Online]. Available: \url{https://doi.org/10.1109/ICDM.2019.00061}
\BIBentrySTDinterwordspacing

\bibitem{oddsketch}
\BIBentryALTinterwordspacing
M.~Mitzenmacher, R.~Pagh, and N.~Pham, ``Efficient estimation for high
  similarities using odd sketches,'' in \emph{23rd International World Wide Web
  Conference, {WWW} '14, Seoul, Republic of Korea, April 7-11, 2014}, 2014, pp.
  109--118. [Online]. Available:
  \url{http://doi.acm.org/10.1145/2566486.2568017}
\BIBentrySTDinterwordspacing

\bibitem{JS_BCS}
\BIBentryALTinterwordspacing
R.~Pratap, I.~Sohony, and R.~Kulkarni, ``Efficient compression technique for
  sparse sets,'' in \emph{Advances in Knowledge Discovery and Data Mining -
  22nd Pacific-Asia Conference, {PAKDD} 2018, Melbourne, VIC, Australia, June
  3-6, 2018, Proceedings, Part {III}}, 2018, pp. 164--176. [Online]. Available:
  \url{https://doi.org/10.1007/978-3-319-93040-4\_14}
\BIBentrySTDinterwordspacing

\bibitem{WeinbergerDLSA09}
\BIBentryALTinterwordspacing
K.~Q. Weinberger, A.~Dasgupta, J.~Langford, A.~J. Smola, and J.~Attenberg,
  ``Feature hashing for large scale multitask learning,'' in \emph{Proceedings
  of the 26th Annual International Conference on Machine Learning, {ICML} 2009,
  Montreal, Quebec, Canada, June 14-18, 2009}, 2009, pp. 1113--1120. [Online].
  Available: \url{http://doi.acm.org/10.1145/1553374.1553516}
\BIBentrySTDinterwordspacing

\bibitem{simhash}
\BIBentryALTinterwordspacing
M.~Charikar, ``Similarity estimation techniques from rounding algorithms,'' in
  \emph{Proceedings on 34th Annual {ACM} Symposium on Theory of Computing, May
  19-21, 2002, Montr{\'{e}}al, Qu{\'{e}}bec, Canada}, 2002, pp. 380--388.
  [Online]. Available: \url{http://doi.acm.org/10.1145/509907.509965}
\BIBentrySTDinterwordspacing

\bibitem{zheng2018feature}
\BIBentryALTinterwordspacing
A.~Zheng and A.~Casari, \emph{Feature Engineering for Machine Learning:
  Principles and Techniques for Data Scientists}.\hskip 1em plus 0.5em minus
  0.4em\relax O'Reilly Media, 2018. [Online]. Available:
  \url{https://books.google.co.in/books?id=sthSDwAAQBAJ}
\BIBentrySTDinterwordspacing

\bibitem{JL83}
\BIBentryALTinterwordspacing
W.~B. Johnson and J.~Lindenstrauss, ``Extensions of lipschitz mappings into a
  hilbert space,'' \emph{Conference in modern analysis and probability (New
  Haven, Conn., 1982), Amer. Math. Soc., Providence, R.I.}, pp. 189--206, 1983.
  [Online]. Available: \url{http://dx.doi.org/10.1016/S0022-0000(03)00025-4}
\BIBentrySTDinterwordspacing

\bibitem{Achlioptas03}
\BIBentryALTinterwordspacing
D.~Achlioptas, ``Database-friendly random projections: Johnson-lindenstrauss
  with binary coins,'' \emph{J. Comput. Syst. Sci.}, vol.~66, no.~4, pp.
  671--687, 2003. [Online]. Available:
  \url{http://dx.doi.org/10.1016/S0022-0000(03)00025-4}
\BIBentrySTDinterwordspacing

\bibitem{LiHC06}
\BIBentryALTinterwordspacing
P.~Li, T.~Hastie, and K.~W. Church, ``Very sparse random projections,'' in
  \emph{Proceedings of the Twelfth {ACM} {SIGKDD} International Conference on
  Knowledge Discovery and Data Mining, Philadelphia, PA, USA, August 20-23,
  2006}, 2006, pp. 287--296. [Online]. Available:
  \url{https://doi.org/10.1145/1150402.1150436}
\BIBentrySTDinterwordspacing

\bibitem{KaneN14}
\BIBentryALTinterwordspacing
D.~M. Kane and J.~Nelson, ``Sparser johnson-lindenstrauss transforms,''
  \emph{J. {ACM}}, vol.~61, no.~1, pp. 4:1--4:23, 2014. [Online]. Available:
  \url{https://doi.org/10.1145/2559902}
\BIBentrySTDinterwordspacing

\bibitem{ScholkopfSM97}
B.~Sch{\"{o}}lkopf, A.~J. Smola, and K.~M{\"{u}}ller, ``Kernel principal
  component analysis,'' in \emph{Artificial Neural Networks - {ICANN} '97, 7th
  International Conference, Lausanne, Switzerland, October 8-10, 1997,
  Proceedings}, 1997, pp. 583--588.

\bibitem{MCA}
J.~Blasius and M.~Greenacre, ``Multiple correspondence analysis and related
  methods,'' \emph{Multiple Correspondence Analysis and Related Methods}, 2006.

\bibitem{IM98}
P.~Indyk and R.~Motwani, ``Approximate nearest neighbors: Towards removing the
  curse of dimensionality,'' in \emph{Proceedings of the Thirtieth Annual {ACM}
  Symposium on the Theory of Computing, Dallas, Texas, USA, May 23-26, 1998},
  1998, pp. 604--613.

\bibitem{BroderCFM98}
\BIBentryALTinterwordspacing
A.~Z. Broder, M.~Charikar, A.~M. Frieze, and M.~Mitzenmacher, ``Min-wise
  independent permutations (extended abstract),'' in \emph{Proceedings of the
  Thirtieth Annual {ACM} Symposium on the Theory of Computing, Dallas, Texas,
  USA, May 23-26, 1998}, 1998, pp. 327--336. [Online]. Available:
  \url{http://doi.acm.org/10.1145/276698.276781}
\BIBentrySTDinterwordspacing

\bibitem{GIM99}
\BIBentryALTinterwordspacing
A.~Gionis, P.~Indyk, and R.~Motwani, ``Similarity search in high dimensions via
  hashing,'' in \emph{VLDB'99, Proceedings of 25th International Conference on
  Very Large Data Bases, September 7-10, 1999, Edinburgh, Scotland, {UK}},
  1999, pp. 518--529. [Online]. Available:
  \url{http://www.vldb.org/conf/1999/P49.pdf}
\BIBentrySTDinterwordspacing

\bibitem{LSI}
S.~Deerwester, S.~T. Dumais, G.~W. Furnas, T.~K. Landauer, and R.~Harshman,
  ``Indexing by latent semantic analysis,'' \emph{JOURNAL OF THE AMERICAN
  SOCIETY FOR INFORMATION SCIENCE}, vol.~41, no.~6, pp. 391--407, 1990.

\bibitem{LDA}
D.~M. Blei, A.~Y. Ng, M.~I. Jordan, and J.~Lafferty, ``Latent dirichlet
  allocation,'' \emph{Journal of Machine Learning Research}, vol.~3, p. 2003,
  2003.

\bibitem{NNMF}
D.~D. Lee and H.~S. Seung, ``Algorithms for non-negative matrix
  factorization.'' in \emph{NIPS}, T.~K. Leen, T.~G. Dietterich, and V.~Tresp,
  Eds., 2000, pp. 556--562.

\bibitem{golinko2019generalized}
E.~Golinko and X.~Zhu, ``Generalized feature embedding for supervised,
  unsupervised, and online learning tasks,'' \emph{Information Systems
  Frontiers}, vol.~21, no.~1, pp. 125--142, 2019.

\bibitem{vanDerMaaten2008}
L.~van~der Maaten and G.~Hinton, ``Visualizing data using {t-SNE},''
  \emph{Journal of Machine Learning Research}, vol.~9, pp. 2579--2605, 2008.

\bibitem{Goodfellow-et-al-2016}
I.~Goodfellow, Y.~Bengio, and A.~Courville, \emph{Deep Learning}.\hskip 1em
  plus 0.5em minus 0.4em\relax MIT Press, 2016,
  \url{http://www.deeplearningbook.org}.

\bibitem{8809826}
X.~Li, M.~Chen, and Q.~Wang, ``Discrimination-aware projected matrix
  factorization,'' \emph{IEEE Transactions on Knowledge and Data Engineering},
  vol.~32, no.~4, pp. 809--814, 2020.

\bibitem{JMLR:v21:17-788}
\BIBentryALTinterwordspacing
X.~Zhang, Q.~Mai, and H.~Zou, ``The maximum separation subspace in sufficient
  dimension reduction with categorical response,'' \emph{Journal of Machine
  Learning Research}, vol.~21, no.~29, pp. 1--36, 2020. [Online]. Available:
  \url{http://jmlr.org/papers/v21/17-788.html}
\BIBentrySTDinterwordspacing

\bibitem{DBLP:journals/tnn/ChenYZLK20}
\BIBentryALTinterwordspacing
X.~Chen, H.~Yang, S.~Zhao, M.~R. Lyu, and I.~King, ``Effective data-aware
  covariance estimator from compressed data,'' \emph{{IEEE} Trans. Neural
  Networks Learn. Syst.}, vol.~31, no.~7, pp. 2441--2454, 2020. [Online].
  Available: \url{https://doi.org/10.1109/TNNLS.2019.2929106}
\BIBentrySTDinterwordspacing

\bibitem{8974600}
Q.~Wang, Z.~Qin, F.~Nie, and X.~Li, ``C2dnda: A deep framework for nonlinear
  dimensionality reduction,'' \emph{IEEE Transactions on Industrial
  Electronics}, vol.~68, no.~2, pp. 1684--1694, 2021.

\bibitem{7155531}
M.~Banerjee and N.~R. Pal, ``Unsupervised feature selection with controlled
  redundancy (ufescor),'' \emph{IEEE Transactions on Knowledge and Data
  Engineering}, vol.~27, no.~12, pp. 3390--3403, 2015.

\bibitem{CormodeDIM03}
G.~Cormode, M.~Datar, P.~Indyk, and S.~Muthukrishnan, ``Comparing data streams
  using hamming norms (how to zero in),'' \emph{{IEEE} Trans. Knowl. Data
  Eng.}, vol.~15, no.~3, pp. 529--540, 2003.

\bibitem{KaneNW10}
D.~M. Kane, J.~Nelson, and D.~P. Woodruff, ``An optimal algorithm for the
  distinct elements problem,'' in \emph{Proceedings of the Twenty-Ninth {ACM}
  {SIGMOD-SIGACT-SIGART} Symposium on Principles of Database Systems, {PODS}
  2010, June 6-11, 2010, Indianapolis, Indiana, {USA}}, 2010, pp. 41--52.

\bibitem{Freivalds1977ProbabilisticMC}
R.~Freivalds, ``Probabilistic machines can use less running time,'' in
  \emph{IFIP Congress}, 1977.

\bibitem{upfal}
M.~Mitzenmacher and E.~Upfal, \emph{Probability and computing - randomized
  algorithms and probabilistic analysis}, 2005.

\bibitem{UCI}
M.~Lichman, ``{UCI} machine learning repository,'' 2013.

\bibitem{Gisette}
I.~Guyon, S.~Gunn, A.~Ben-Hur, and G.~Dror, ``Result analysis of the nips 2003
  feature selection challenge,'' in \emph{Advances in Neural Information
  Processing Systems 17}, 2005, pp. 545--552.

\bibitem{DeliciousMIL}
\BIBentryALTinterwordspacing
H.~Soleimani and D.~J. Miller, ``Semi-supervised multi-label topic models for
  document classification and sentence labeling,'' in \emph{Proceedings of the
  25th {ACM} International Conference on Information and Knowledge Management,
  {CIKM} 2016, Indianapolis, IN, USA, October 24-28, 2016}, 2016, pp. 105--114.
  [Online]. Available: \url{https://doi.org/10.1145/2983323.2983752}
\BIBentrySTDinterwordspacing

\bibitem{genomics20171}
X.~Genomics, ``1.3 million brain cells from e18 mice,'' \emph{CC BY}, vol.~4,
  2017.

\bibitem{SSD}
G.~Cormode, M.~Datar, P.~Indyk, and S.~Muthukrishnan, ``Comparing data streams
  using hamming norms (how to zero in),'' \emph{{IEEE} Trans. Knowl. Data
  Eng.}, vol.~15, no.~3, pp. 529--540, 2003.

\bibitem{Kingma2014}
D.~P. Kingma and M.~Welling, ``{Auto-Encoding Variational Bayes},'' in
  \emph{2nd International Conference on Learning Representations, {ICLR} 2014,
  Banff, AB, Canada, April 14-16, 2014, Conference Track Proceedings}, 2014.

\bibitem{Sulc2015DimensionalityRO}
Z.~Sulc and H.~Rezankov{\'a}, ``Dimensionality reduction of categorical data:
  Comparison of hca and catpca approaches,'' 2015.

\bibitem{mcdiarmid_1989}
C.~McDiarmid, \emph{On the method of bounded differences}, ser. London
  Mathematical Society Lecture Note Series.\hskip 1em plus 0.5em minus
  0.4em\relax Cambridge University Press, 1989, p. 148–188.

\bibitem{kmode}
\BIBentryALTinterwordspacing
Z.~Huang, ``Extensions to the k-means algorithm for clustering large data sets
  with categorical values,'' \emph{Data Mining and Knowledge Discovery},
  vol.~2, no.~3, pp. 283--304, 1998. [Online]. Available:
  \url{https://doi.org/10.1023/A:1009769707641}
\BIBentrySTDinterwordspacing

\bibitem{CORMODE200558}
\BIBentryALTinterwordspacing
G.~Cormode and S.~Muthukrishnan, ``An improved data stream summary: the
  count-min sketch and its applications,'' \emph{J. Algorithms}, vol.~55,
  no.~1, pp. 58--75, 2005. [Online]. Available:
  \url{https://doi.org/10.1016/j.jalgor.2003.12.001}
\BIBentrySTDinterwordspacing

\bibitem{CHARIKAR20043}
M.~Charikar, K.~Chen, and M.~Farach-Colton, ``Finding frequent items in data
  streams,'' \emph{Theoretical Computer Science}, vol. 312, no.~1, pp. 3 -- 15,
  2004, automata, Languages and Programming.

\bibitem{Indyk06}
\BIBentryALTinterwordspacing
P.~Indyk, ``Stable distributions, pseudorandom generators, embeddings, and data
  stream computation,'' \emph{J. {ACM}}, vol.~53, no.~3, pp. 307--323, 2006.
  [Online]. Available: \url{http://doi.acm.org/10.1145/1147954.1147955}
\BIBentrySTDinterwordspacing

\bibitem{FLDA}
R.~Fisher, ``The statistical utilization of multiple measurements,''
  \emph{Annals of Eugenics}, vol.~8, pp. 376--386, 1938.

\bibitem{10.5555/2987189.2987190}
G.~E. Hinton and R.~S. Zemel, ``Autoencoders, minimum description length and
  helmholtz free energy,'' in \emph{Proceedings of the 6th International
  Conference on Neural Information Processing Systems}, ser. NIPS'93.\hskip 1em
  plus 0.5em minus 0.4em\relax San Francisco, CA, USA: Morgan Kaufmann
  Publishers Inc., 1993, p. 3–10.

\bibitem{NIPS1998_226d1f15}
\BIBentryALTinterwordspacing
S.~Mika, B.~Sch\"{o}lkopf, A.~Smola, K.-R. M\"{u}ller, M.~Scholz, and
  G.~R\"{a}tsch, ``Kernel pca and de-noising in feature spaces,'' in
  \emph{Advances in Neural Information Processing Systems}, M.~Kearns,
  S.~Solla, and D.~Cohn, Eds., vol.~11.\hskip 1em plus 0.5em minus 0.4em\relax
  MIT Press, 1999. [Online]. Available:
  \url{https://proceedings.neurips.cc/paper/1998/file/226d1f15ecd35f784d2a20c3ecf56d7f-Paper.pdf}
\BIBentrySTDinterwordspacing

\bibitem{tenenbaum_global_2000}
J.~B. Tenenbaum, V.~de~Silva, and J.~C. Langford, ``A global geometric
  framework for nonlinear dimensionality reduction,'' \emph{Science}, vol. 290,
  no. 5500, p. 2319, 2000.

\bibitem{kohonen-self-organized-formation-1982}
\BIBentryALTinterwordspacing
T.~Kohonen, ``Self-organized formation of topologically correct feature maps,''
  \emph{Biological Cybernetics}, vol.~43, no.~1, pp. 59--69, Jan. 1982.
  [Online]. Available: \url{http://dx.doi.org/10.1007/BF00337288}
\BIBentrySTDinterwordspacing

\end{thebibliography}

%








\begin{IEEEbiography}[{\includegraphics[width=1in,height=1.25in,clip,keepaspectratio]{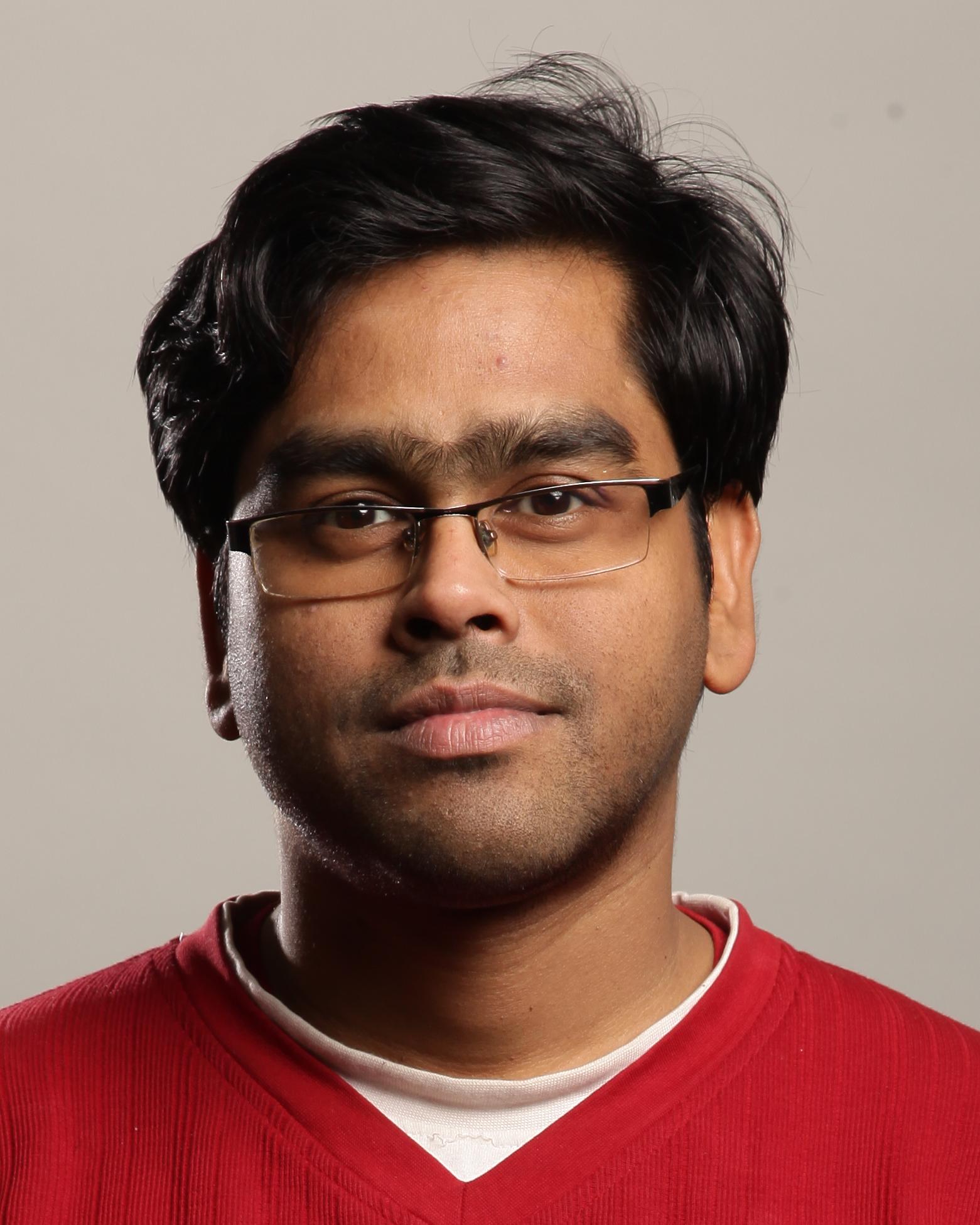}}]{Debajyoti~Bera}
received his B.Tech. in Computer Science and Engineering in 2002 at Indian Institute of Technology (IIT), Kanpur, India and his Ph.D. degree in Computer Science from Boston University, Massachusetts, USA in 2010. Since 2010 he is an assistant professor at Indraprastha Institute of Information Technology, (IIIT-Delhi), New Delhi, India.
His research interests include quantum computing, randomized algorithms, and engineering algorithms for networks, data mining, and information security.
\end{IEEEbiography}
\vspace{-4mm}

\begin{IEEEbiography}[{\includegraphics[width=1in,height=1.25in,clip,keepaspectratio]{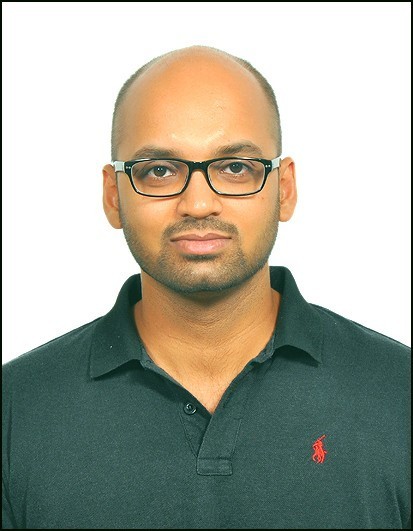}}]{Rameshwar Pratap}
 has earned Ph.D in Theoretical Computer Science in 2014  from Chennai Mathematical Institute (CMI). Earlier, he completed Masters in Computer Application (MCA) from  Jawaharlal Nehru University and   BSc in Mathematics, Physics, and Computer Science from University of Allahabad. Post Ph.D he has worked  TCS Innovation Labs (New Delhi, India),  and Wipro AI-Research (Bangalore, India). Since 2019 he is working as an assistant professor at School of Computing and Electrical Engineering (SCEE), IIT Mandi. His research interests include algorithms for dimensionality reduction,  robust sampling, and algorithmic fairness. 
 \end{IEEEbiography}
 
 \begin{IEEEbiography}[{\includegraphics[width=1in,height=1.25in,clip,keepaspectratio]{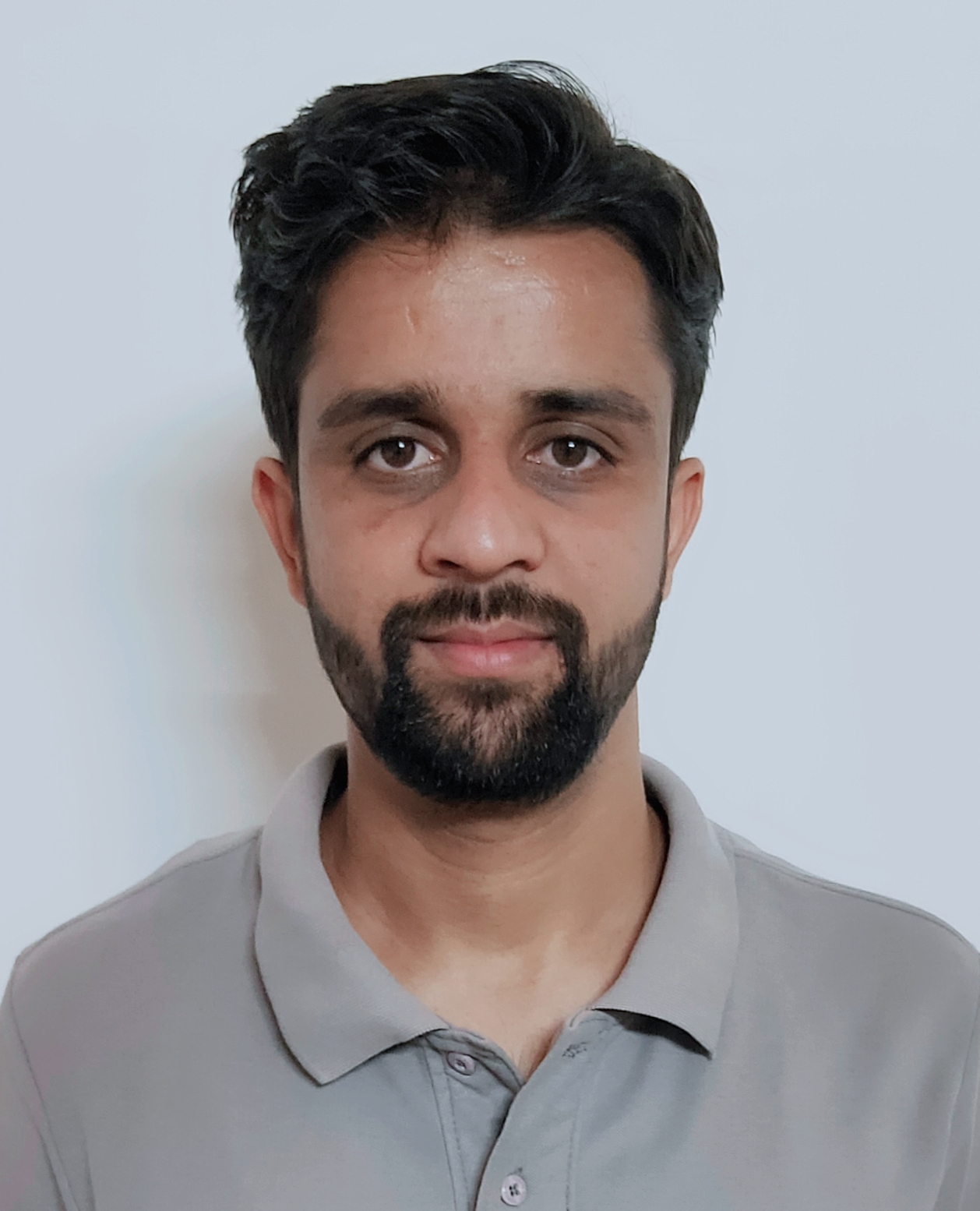}}]{Bhisham Dev Verma}
 is pursuing Ph.D from IIT Mandi. He has done his Masters in Applied Mathematics from IIT Mandi and BSc in Mathematics, Physics, and Chemistry from Himachal Pradesh University. His research interest includes data mining, algorithms for dimension reduction, optimization and machine learning.
 \end{IEEEbiography}

\newpage
\clearpage
\appendices

\section{Analysis of one-hot encoding + binary compression}\label{appendix:OHE+BS}

Let $x$ and $y$ be two $n$-dimensional categorical vectors with sparsity at most $\sigma$; $c$ will denote the maximum number of values any attribute can take. Let $x'$ and $y'$ be the one-hot encodings of $x$ and $y$, respectively. Further, let $x''$ and $y''$ denote the compression of $x'$ and $y'$, respectively, using BinSketch~\cite{ICDM} which is the state-of-the-art dimensionality reduction for binary vector using Hamming distance.

Observe that the sparsity of $x'$ is same as that of $x$ and a similar claim holds for $y'$ and $y$. However, $HD(x',y')$ does not hold a monotonic relationship with $HD(x,y)$. 
It is easy to show that $HD(x,y) \le HD(x',y') \le 2HD(x,y)$. Therefore, 
\begin{equation}
    |HD(x,y) - HD(x',y')| \le HD(x,y) \le 2\sigma. \label{eqn:2}
\end{equation}
    
We need the following lemma that was used to analyse BinSketch~\cite[Lemma~12,Appendix~A]{ICDM}.
\begin{lem}
Suppose we compress two $n'$-dimensional binary vectors $x'$ and $y'$ with sparsity at most $\sigma$ to $g$-dimensional binary sketches, denotes $x''$ and $y''$ respectively, by following an algorithm proposed in the BinSketch work. If $g$ is set to $\sigma\sqrt{\tfrac{\sigma}{2}\ln \frac{6}{\delta}}$ for any $\delta \in (0,1)$, then the following holds with probability at least $1-\delta$.
$$|HD(x',y') - HD(x'',y'')| \le 6 \sqrt{\tfrac{\sigma}{2} \ln \tfrac{6}{\delta}}.$$
\end{lem}

Combining the above inequality with that in Equation~\ref{eqn:2} gives us
$$|HD(x,y)-HD(x'',y'')| \le 2\sigma + 6\sqrt{\tfrac{\sigma}{2} \ln \tfrac{6}{\delta}} \le 2\sigma\sqrt{\ln \tfrac{2}{\delta}}$$
if we set the reduced dimension to $\sigma\sqrt{\tfrac{\sigma}{2}\ln \frac{6}{\delta}}$.

This bound is worse compared to that of \fsketch where we able to prove an accuracy of $\Theta(\sqrt{\sigma\ln \tfrac{2}{\delta}})$ using reduced dimension value of $4\sigma$ (see Lemma~\ref{lem:hconcentrationlemma}).

\section{Proofs from Section~\ref{subsec:analysis}}\label{appendix:subsec:analysis}

\concentrationlemma*

\begin{proof}
Fix any $R$ and $x,y$; the rest of the proof applies to any $R$, and therefore, holds for a random $R$ as well. Define a vector $z \in \{0,\pm 1,\ldots,\pm c\}^n$ in which $z_i=(x_i - y_i)$; the number of non-zero entries of $z$ are at most $2\sigma$ since the number of non-zero entries of $x$ and $y$ are at most $\sigma$. Let $J_0$ be the set of coordinates from $\{1, \ldots, n\}$ at which $z$ is 0, and let $J_1$ be the set of the rest of the coordinates; from above, $J_1 \le 2\sigma$.

Define the event $E_j$ as ``$[\phi_j(x)\not=\phi_j(y)]$''. Note that $f$ can be written as a sum of indicator random variables, $\sum_j I(E_j)$, and we would like to prove that $f$ is almost always close to $f^*=\E[f]$.

Observe that $\phi_j(x)=\phi_j(y)$ iff $\sum_{i \in \rho^{-1}(j)} z_i \cdot r_i = 0 \mod{p}$ iff $\sum_{i \in \rho^{-1}(j) \cap J_1} z_i \cdot r_i = 0 \mod{p}$. In other words, $\rho(i)$ could be set to anything for $i \in J_0$ without any effect on the event $E_j$; hence, we will assume that the mapping $\rho$ is defined as a random mapping only for $i \in J_1$, and further for the ease of analysis, we will denote them as $\rho(i_1), \rho(i_2), \ldots, \rho(i_{2\sigma})$ (if $|J_1| < 2\sigma$ then move a few coordinates from $J_0$ to $J_1$ without any loss of correctness).

To prove the concentration bound we will employ martingales. Consider the sequence of these random variables $\rho'=\rho(i_1), \rho(i_2), \ldots, \rho(i_{2\sigma})$ -- these are independent. Define a function $g(\rho')$ of these random variables as a sum of indicator random variables as stated below (note that $R$ and $\rho(i)$, for $i \in J_0$, are fixed at this point)
\begin{align*}
    & g(\rho(i_1), \rho(i_2), \ldots \rho(i_{2\sigma}))\\
    & = \sum_j I\left( \sum_{i \in \rho^{-1}(j) \cap J_1} z_i \cdot r_i \not= 0 \mod{p} \right) \\
    & = \sum_j I(E_j) = f
\end{align*}

Now consider an arbitrary $t\in \{1, \ldots, 2\sigma\}$ and let $q=\rho(i_t)$; observe that $z_{i_t}$ influences only $E_q$. Choose an arbitrary value $q' \in \{1, \ldots, d\}$ that is different from $q$. Observe that, if $\rho$ is modified only by setting $\rho(i_t)=q'$ then we claim that ``bounded difference holds''.

\begin{prop}
$|~ g(\rho(i_1), \ldots, \rho(i_{t-1}), q, \ldots, \rho(i_{2\sigma})) - g(\rho(i_1), \ldots, \rho(i_{t-1}), q', \ldots, \rho(i_{2\sigma})) ~| \le 2$.
\end{prop}
The proposition holds since the only effects of the change of $\rho(i_t)$ from $q$ to $q'$ are seen in $E_q$ and $E_{q'}$ (earlier $E_q$ depended upon $z_{i_t}$ that now changes to $E_{q'}$ being depended upon $z_{i_t}$). Since $g()$ obeys bounded difference, therefore, we can apply McDiarmid's inequality~\cite[Ch~17]{upfal},~\cite{mcdiarmid_1989}.
\begin{thm}[McDiarmid's inequality\label{thm:McDiarmid}]
Consider independent random variables $X_1,\ldots , X_m \in \mathcal{X}$, and a mapping $ f:\mathcal{X}^m \rightarrow \R$ which   for all $i$ and for all $x_1,\ldots x_m, {x_i}'$ satisfies the property:
$|f(x_1,\ldots,x_i,\ldots,x_m)-f(x_1,\ldots,{x_i}',\ldots,x_m)|\leq c_i$, where $x_1,\ldots x_m, {x_i}'$ are possible values for the input variables of the function $f$. Then,
    \begin{align*}
	\Pr\Big[ \big| \E[f(X_1,\ldots , X_m) - f(X_1,\ldots , X_m)]\big|  \geq \varepsilon \Big] \\
	\leq 2\exp \left(\frac{-2\varepsilon^2}{\sum_{i=1}^m c_i^2} \right).
    \end{align*}
\end{thm}

This inequality implies that, for every $x,y,R$,
$$\Pr_{\rho} \Big[ \big| \E[f] - f \big| \ge \alpha \Big] \le 2\exp\left( -\frac{2\alpha^2}{(2\sigma)2^2} \right) = \exp \left( - \frac{\alpha^2}{4\sigma} \right).$$
Hence, the lemma is proved.
\end{proof}

\upperboundf*

\begin{proof}
    Since $f^*=dP(1-D^h)=dP - dPD^h$, if $f \ge dP$ then $|f-f^*| \ge dPD^h$. 
    \begin{align*}
	\Pr[f \ge dP] & \le \Pr[|f-f^*| \ge dPD^h]\\
		      & \le 2 \exp(-\tfrac{d^2 P^2 D^{2h}}{4\sigma}) \tag{using Lemma~\ref{lem:concentration}}\\
		      & = 2 \exp(-\tfrac{P^2}{4\sigma} d^2(1-\tfrac{1}{d})^{2h})\\
		      & \le 2 \exp(-\tfrac{P^2}{4\sigma} (d-h)^2) \tag{$\because~(1-\tfrac{1}{d})^h \ge 1-\tfrac{h}{d}$}\\
		      & \le 2 \exp(-P^2\sigma) \tag{$\because$ $\tfrac{(d-h)^2}{4\sigma} \ge \sigma$}
    \end{align*}
    Here we have used the fact that $h \le 2\sigma$ which, along with the setting $d=4\sigma$, implies that $(d-h) \ge 2\sigma$.
\end{proof}

\hconcentrationlemma*

\begin{proof}
    Denote $|\hat{h}-h|$ by $\Delta h$ and let $\alpha=\sqrt{d \ln
    \tfrac{2}{\delta}}$. We will prove that $\Delta h < \tfrac{32}{P}\sqrt{\sigma \ln \tfrac{2}{\delta}}$ for the case $|f-f^*| \le \alpha$ which, by
    Lemma~\ref{lem:concentration}, happens
    with probability at least $(1-2\exp{(-\tfrac{\alpha^2}{4\sigma})})=1-\delta$.

    First we make a few technical observations all of which are based on
    standard inequalities of binomial series and logarithmic functions.
    It will be helpful to remember that $D=1-1/d \in (0,1)$.
    \begin{obs}\label{obs:primep}
	For reasonable values of $\sigma$, and
	reasonable values of $\delta$, almost all
	primes satisfy the bound $P \ge \tfrac{4}{\sqrt{\sigma}}\sqrt{\ln
	\tfrac{2}{\delta}}$. We will assume this inequality to hold without loss
	of generality~\footnote{If the reader is wondering why we are not
	proving this fact, it may be observed that this relationship does not
	hold for small values of $\sigma$, e.g., $\sigma=16, \delta=0.01$.}.
    \end{obs}
    For example, $p=2$ is sufficient for $\sigma \approx 1000$ and $\delta
    \approx 0.001$ (remember that $P=1-\tfrac{1}{p}$). Furthermore, observe that
    $P$ is an increasing function of $p$, and the right hand side is a
    decreasing function of $\sigma$, increasing with decreasing delta but at an
    extremely slow logarithmic rate.
    
    \begin{obs}\label{obs:dp-alpha}
	$\tfrac{dP}{\alpha} > 4$ can be assumed without loss of generality. This holds since the left hand side is
	$\tfrac{dP}{\sqrt{d}\sqrt{ln(2/\delta)}} =
	\tfrac{P\sqrt{d}}{\sqrt{\ln(2/\delta)}} \ge
	\tfrac{4\sqrt{d}}{\sqrt{\sigma}}$ (by Observation~\ref{obs:primep})
	which is at least $4$.
    \end{obs}

    \begin{obs}\label{obs:f_less_than_dP}
	Based on the above assumptions, $f < dP$.
    \end{obs}
    \begin{proof}[Proof of Observation] We will prove that $\sqrt{d \ln \tfrac{2}{\delta}} < dPD^h$. Since $|f-f^*| \le \sqrt{d \ln \tfrac{2}{\delta}}$ and $f^*=dP(1-D^h)$, it follows that $f \le f^* + \sqrt{d \ln \tfrac{2}{\delta}} < dP$.

    \begin{align*}
	\sqrt{d}PD^h & = \frac{dPD^h}{\sqrt{d}} \ge \frac{P}{\sqrt{d}}d(1-\tfrac{1}{d})^h \ge \frac{P}{\sqrt{d}}d(1-\tfrac{h}{d})\\
		     & = \frac{P}{\sqrt{d}}(d-h) \ge \frac{P}{\sqrt{d}}\frac{d}{2} \tag{$\because$ $h \le 2\sigma$, $d-h \ge 2\sigma = \tfrac{d}{2}$}\\
		     & = P \sqrt{\sigma} \ge 4\sqrt{\ln \tfrac{2}{\delta}} \tag{Observation~\ref{obs:primep}}
    \end{align*}
    which proves the claim stated at the beginning of the proof.
    \end{proof}

    Based on this observation, $\hat{h}$ is calculated as $\ln\left( 1 - \tfrac{f}{dP} \right)/\ln D$ (see Definition~\ref{defn:estimator}). 
    Thus, we get $D^{\hat{h}} = 1 - \tfrac{f}{dP}$.
    Further, from Equation~\ref{eqn:fstar} we get $D^h = 1-\tfrac{f^*}{dP}$.

    \begin{obs}\label{obs:appendix-1} $D^h \ge D^{2\sigma} \ge \tfrac{9}{16}$. This is since $h \le 2\sigma$ and 
	$D^\sigma = (1-\tfrac{1}{d})^\sigma \ge 1 - \tfrac{\sigma}{d} =
	\tfrac{3}{4}$.
    \end{obs}

    \begin{restatable}{obs}{obsappendix}\label{obs:appendix-2}
	$D^{\hat{h}} > \tfrac{5}{16}$.
    \end{restatable}
	This is not so straight forward as
	Observation~\ref{obs:appendix-1} since $\hat{h}$ is calculated using a
	formula and is not guaranteed, ab initio, to be upper bounded by
	$2\sigma$.
	\begin{proof}[Proof of Observation]
	    We will prove that $\tfrac{f}{dP} < \tfrac{11}{16}$ which will imply that $D^{\hat{h}} = 1 - \tfrac{f}{dP} > \tfrac{5}{16}$.
	   
	    For the proof of the lemma we have considered the case that $f \le f^* + \alpha$. 
	    Therefore, $\tfrac{f}{dP} \le \tfrac{f^*}{dP} + \tfrac{\alpha}{dP}$.
	    Substituting the value of $f^*=dP(1-D^h)$ from Equation~\ref{eqn:fstar} and
	    using Observation~\ref{obs:appendix-1} we get the bound $\tfrac{f}{dP} \le
	    \tfrac{7}{16} + \tfrac{\alpha}{dP}$. We can further simplify the bound using Observation~\ref{obs:dp-alpha}:
    $$\tfrac{f}{dP} \le \tfrac{7}{16} + \tfrac{\alpha}{dP} \le \tfrac{7}{16} +
    \tfrac{1}{4} < \tfrac{11}{16}, \mbox{ validating the observation.}$$

	\end{proof}

    Now we get into the main proof which proceeds by considering two possible
    cases.
    
    {\bf (Case $\hat{h} \ge h$, i.e., $\Delta h=\hat{h}-h$:)} 
    We start with the identity $D^h - D^{\hat{h}} = \tfrac{f-f^*}{dP}$.

    Notice that the RHS is bounded from the above by $\tfrac{\alpha}{dP}$ and
    the LHS can bounded from the below as
    $$D^h - D^{\hat{h}} = D^h(1-D^{\Delta h}) > \tfrac{9}{16}(1-D^{\Delta h})$$
    where we have used Observation~\ref{obs:appendix-1}. Combining these facts
    we get $\tfrac{\alpha}{dP} > \tfrac{9}{16}(1-D^{\Delta h})$.

    {\bf (Case $h \ge \hat{h}$, i.e., $\Delta h = h - \hat{h}$:)} In a similar
    manner, we start with the identity $D^{\hat{h}} - D^h = \tfrac{f^* -
    f}{dP}$ in which the RHS we bound again from the above by
    $\tfrac{\alpha}{dP}$ and the LHS is treated similarly (but now using
    Observation~\ref{obs:appendix-2}).
    $$D^{\hat{h}} - D^h = D^{\hat{h}}(1-D^{\Delta h}) > \tfrac{5}{16}
    (1-D^{\Delta h})$$ and then, $\tfrac{\alpha}{dP} >
    \tfrac{5}{16}(1-D^{\Delta h})$.

    So in both the cases we show that $\tfrac{\alpha}{dP} >
    \tfrac{5}{16}(1-D^{\Delta h})$. Our desired bound on $\Delta h$ can now be
    obtained.
    \begin{align*}
	\Delta h \ln D & \ge \ln\left( 1 - \tfrac{16}{5}\tfrac{\alpha}{dP} \right)
			 \ge -\tfrac{16\alpha}{5dP}/(1-\tfrac{16\alpha}{5dP}) =
	-\tfrac{16\alpha}{5dP-16\alpha}\\
	& \mbox{ (using the inequality $\ln(1+x) \ge
	\tfrac{x}{1+x}$ for $x > -1$)} \\
	\therefore \Delta h & \le \frac{1}{\ln\tfrac{1}{D}} \frac{16\alpha}{5dP -
	16\alpha} \le \frac{16\alpha d}{5dP - 16\alpha}\\
	& \mbox{ (it is easy to show that $\ln \tfrac{1}{D} =
	\ln\tfrac{1}{1-1/d} \ge 1/d$)}\\
	& = \frac{\tfrac{16}{5}d}{\tfrac{dP}{\alpha} - \tfrac{16}{5}} \\
	& < \frac{\tfrac{16}{5}d}{\tfrac{dP}{5\alpha}} 
	\mbox{ (using Observation~\ref{obs:dp-alpha}, $\tfrac{dP}{\alpha}-\tfrac{16}{5} >\tfrac{dP}{5\alpha}$)}\\
	& = \frac{16\alpha}{P} = \frac{16}{P}\sqrt{d \ln \tfrac{2}{\delta}} =
	\frac{32}{P}\sqrt{\sigma \ln \tfrac{2}{\delta}}
    \end{align*}
\end{proof}

\hconcentrationlemmatight*

\begin{proof}[Proof of (a) $f < dP$ with high probability]
    Following the steps of the proof of Lemma~\ref{lem:upper-bound-f},
    \begin{align*}
	\Pr[f \ge dP] & \le 2\exp(-\tfrac{d^2 P^2 D^{2h}}{4\sigma}) \\
		      & \le 2\exp(-P^2\tfrac{(d-h)^2}{4\sigma})
    \end{align*}
    Let $L$ denote $\sqrt{\ln\tfrac{2}{\delta}}$; note that $L > 1$. Now, $d=16L\sqrt{\sigma}$ and $h \le \sqrt{\sigma}$. So, $d-h \ge 15L\sqrt{\sigma} > 15\sqrt{\sigma}$ and, therefore, $\tfrac{(d-h)^2}{\sigma} > 225$. Using this bound in the equation above, we can upper bound the right-hand side as $2\exp(-225(1-\tfrac{1}{p})^2/4)$ which is a decreasing function of $p$, the lowest (for $p=2$) being $2\exp(-225/4*4) \approx 10^{-6}$.
\end{proof}

\begin{proof}[Proof of (b) a better estimator of $h$]
    The proof is almost exactly same as that of
    Lemma~\ref{lem:hconcentrationlemma}, with only a few differences. We set
    $\alpha=d/8$ where $d=16\sqrt{\sigma \ln \tfrac{2}{\delta}}$. Incidentally,
    the value of $\alpha$ remains the same in terms of $\sigma$ ( $\alpha=\sqrt{4\sigma
    \ln \tfrac{2}{\delta}}$). Thus, the
    probability of error remains same as before;
    $$2 \exp{(-\tfrac{d^2}{64 \cdot 4 \sigma})} = \delta.$$

    Observation~\ref{obs:primep} is true without any doubt.
    $\tfrac{dP}{\alpha} = 8P$ which is greater than 4 for any prime number; so
    Observation~\ref{obs:dp-alpha} is true in this scenario.

    Observation~\ref{obs:f_less_than_dP} requires a new proof. Following the steps of the above proof of Observation~\ref{obs:f_less_than_dP}, it suffices to prove that $dPD^h > \tfrac{d}{8}$.
    \begin{align*}
	PD^h & = P(1-\tfrac{1}{d})^h \ge P(1-\tfrac{h}{d}) \\
	     & =P(\tfrac{d-h}{d}) \ge P\tfrac{15L\sqrt{\sigma}}{16L\sqrt{\sigma}} = P\tfrac{15}{16} > \tfrac{1}{2}\tfrac{15}{16} > \tfrac{1}{8}
    \end{align*}

    Observation~\ref{obs:appendix-1} is now tighter since $D^h \ge D^{\sqrt{\sigma}}
    = (1-\tfrac{1}{d})^{\sqrt{\sigma}} \ge 1-\tfrac{\sqrt{\sigma}}{d} = 1 -
    \tfrac{1}{16\sqrt{\ln 2/\delta}} \ge \tfrac{3}{4}$ for reasonable values of
    $\delta$.
    Similarly Observation~\ref{obs:appendix-2} is also tighter (it relies on only the above
    observations) since $\tfrac{f^*}{dP} = 1 - D^h \le 1-\tfrac{3}{4}$ and $\tfrac{\alpha}{dP} < \tfrac{1}{4}$; we get $D^{\hat{h}} > \tfrac{1}{2}$.

    Following similar steps as above, for the case $\hat{h} \ge h$, we get $\tfrac{\alpha}{dP} > \tfrac{3}{4}(1-D^{\Delta h})$ and for the case $\hat{h} < h$, we get $\tfrac{\alpha}{dP} > \tfrac{1}{2}(1-D^{\Delta h})$ leading to the common condition that $\tfrac{\alpha}{dP} > \tfrac{1}{2}(1-D^{\Delta h})$.

    The final thing to calculate is the bound on $\Delta h$.
    \begin{align*}
	\Delta h \ln D & \ge \ln\left( 1 - \tfrac{2\alpha}{dP} \right)
			 \ge -\tfrac{2\alpha}{dP}/(1-\tfrac{2\alpha}{dP}) =
	-\tfrac{2\alpha}{dP-2\alpha}\\
	& \mbox{ (using the inequality $\ln(1+x) \ge
	\tfrac{x}{1+x}$ for $x > -1$)} \\
	\therefore \Delta h & \le \frac{1}{\ln\tfrac{1}{D}} \frac{2\alpha}{dP -
	2\alpha} \le \frac{2\alpha d}{dP - 2\alpha}\\
	& \mbox{ (it is easy to show that $\ln \tfrac{1}{D} =
	\ln\tfrac{1}{1-1/d} \ge 1/d$)}\\
	& = \frac{2d}{\tfrac{dP}{\alpha} - 2} \\
	& < \frac{2d}{\tfrac{dP}{2\alpha}} 
	\mbox{ (using Observation~\ref{obs:dp-alpha}, $\tfrac{dP}{\alpha}-2 >\tfrac{dP}{2\alpha}$)}\\
	& = \frac{4\alpha}{P} = \frac{4}{P}\sqrt{4\sigma \ln \tfrac{2}{\delta}} =
	\frac{8}{P}\sqrt{\sigma \ln \tfrac{2}{\delta}}
    \end{align*}
\end{proof}

\section{Complexity analysis of \fsketch}\label{subsec:appendix-complexity}
There
are two major operations with respect to \fsketch --- construction of sketches
and estimation of Hamming distance from two sketches. We will discuss their time and space requirements. There are efficient representations of sparse data vectors, but for the sake of simplicity we assume full-size arrays to store vectors; similarly we assume simple dictionaries for storing the interval variables $\rho,R$ by \fsketch. While it may be possible to reduce the number of random bits by employing $k$-wise independent bits and mappings, we left it out of the scope of this work and for future exploration.
\begin{enumerate}
    \item{Construction:} Sketches are constructed by the \fsketch algorithm which does a linear pass over the input vector, maps every non-zero attribute to some entry of the sketch vector and then updates that corresponding entry. The time to process one data vector becomes $\Theta(n) + O(\sigma \cdot poly(\log p))$ which is $O(n)$ for constant $p$. The interval variables, $\rho,R,p$, require space $\Theta(n \log d)$, $\Theta(n \log p)$ and $\Theta(\log p)$, respectively, which is almost $O(n)$ if $\sigma \ll n$. Furthermore, $\rho$ and $R$, that can consume bulk of this space, can be freed once the sketch construction phase is over. A sketch itself consumes $\Theta(d \log p)$ space.
    \item{Estimation:} There is no additional space requirement for estimating the Hamming distance of a pair of points from their sketches. The estimator scans both the sketches and computes their Hamming distance; finally it computes an estimate by using Definition~\ref{defn:estimator}. The running time is $O(d \log p)$.
\end{enumerate}

\section{Proofs from Section~\ref{subsec:complexity}}\label{appendix:subsec:complexity}

\sparsitylemma*

\begin{proof}
The lemma can be proved by treating it as a balls-and-bins problem. Imagine
    throwing $\sigma$ balls (treat them as the non-zero attributes of $x$) into
    $d$ bins (treat them as the sketch cells) independently and uniformly at
    random. If the $j$th-bin remains empty then $\phi_j(x)$ must be zero (the
    converse is not true). Therefore, the expected number of non-zero cells in the
    sketch is upper bounded by the expected number of empty bins, which can be
    easily shown to be $d[1-(1-\tfrac{1}{d})^\sigma]$. Using the stated value of
    $d$, this expression can further be upper bounded.
    $$d[1-(1-\tfrac{1}{d})^\sigma] \le d[1-(1-\tfrac{\sigma}{d})] =
    \tfrac{d}{4}$$
    Furthermore, let $NZ$ denote the number of non-zero entries in $\phi(x)$. We
    derived above $\E[NZ] \le \tfrac{d}{4}$. Markov inequality can help in upper
    bounding the probability that $\phi(x)$ contains many non-zero entries.
    $$\Pr[NZ \ge \tfrac{d}{2}] \le \E[NZ]/\tfrac{d}{2} \le \tfrac{1}{2}$$
\end{proof}

\section{ Reproducibility details}\label{sec:Reproducibility_details}
\subsection{Baseline implementations }\label{sec:Reproducibility_baseline}

\begin{enumerate}
 \item  We implemented the {feature hashing (FH)}~\cite{WeinbergerDLSA09},   {SimHash (SH)}\cite{simhash}, {Sketching via Stable Distribution (SSD)}\cite{SSD} and {One Hot Encoding (OHE)}\cite{ICDM} algorithms on own own; we have made these implementations publicly available~\footnote{\url{https://github.com/Anonymus135/F-Sketch}}.
 \item For  {Kendall rank correlation coefficient}\cite{kendall1938measure} we used the implementation provided by \texttt{pandas data frame}~\footnote{\url{https://pandas.pydata.org/pandas-docs/stable/reference/api/pandas.DataFrame.corr.html}}. 
 \item For {Latent Semantic Analysis (LSA)}~\cite{LSI}, {Latent Dirichlet Allocation (LDA)}\cite{LDA}, {Non-negative Matrix Factorisation (NNMF)}\cite{NNMF}, and  vanilla {Principal component analysis (PCA)}, we used their implementations available in the \texttt{sklearn.decomposition} library~\footnote{\url{https://scikit-learn.org/stable/modules/classes.html\#module-sklearn.decomposition}}.
 \item For {Multiple Correspondence Analysis (MCA)}~\cite{MCA}, we used a Python library~\footnote{\url{https://pypi.org/project/mca/}}.
 \item { For HCA~\cite{Sulc2015DimensionalityRO}, we performed hierarchical clustering~\footnote{\url{https://scikit-learn.org/stable/modules/generated/sklearn.cluster.AgglomerativeClustering.html}} over the features in which we set the number of clusters to the value of reduced dimension. We then randomly sampled one feature from each of the clusters, and considered the data points restricted to the sampled features.
 \item For CATPCA~\cite{Sulc2015DimensionalityRO}, we
used an R package~\footnote{\url{https://rdrr.io/rforge/Gifi/man/princals.html}}.}

 \end{enumerate}
It should be noted that PCA, MCA and LSA cannot reduce the dimension beyond the number of data points.

\subsection{Reproducibility details for clustering task}\label{appendix:clustering}
    We first generated the ground truth clustering results on the datasets using $k$-mode~\cite{kmode} (we used a Python library~\footnote{\url{https://pypi.org/project/kmodes/}}).
    
    We then compressed the datasets using the baselines. Of them, {feature hashing}~\cite{WeinbergerDLSA09}, {SimHash}~\cite{simhash}, and {Kendall rank correlation coefficient}\cite{kendall1938measure} generate integer/discrete valued  sketches on which we can define a Hamming distance. Therefore we use the $k$-mode algorithm on compressed datasets. On the other hand, {Latent Semantic Analysis (LSA)}~\cite{LSI}, {Latent Dirichlet Allocation (LDA)}~\cite{LDA}, {Non-negative Matrix Factorisation (NNMF)}~\cite{NNMF},  {Principal component analysis (PCA)}, and {Multiple Correspondence Analysis (MCA)}~\cite{MCA} generate real-valued sketches. For these we used the $k$-means algorithm (available in the \texttt{sklearn} library~\footnote{\url{https://scikit-learn.org/stable/modules/classes.html\#module-sklearn.cluster}}) on the compressed datasets. We set \texttt{random\_state = 42} for both $k$-mode and $k$-means.
    
    We evaluated the clustering outputs using {\em purity index}. Let $m$ be the number of data points and $\Omega=\{\omega_1, \omega_2,\ldots, \omega_k\}$ be a set of 
   clusters obtained on the original data. Further, let $\mathcal{C}=\{c_1, c_2,\ldots, c_k\}$ be a set of 
   clusters obtained on reduced dimensional data. Then the \textit{purity index} of the clusters $\mathcal{C}$ is defined as
 $$
   \textit{purity index}(\Omega, \mathcal{C})=\frac{1}{m}\sum_{i=1}^k \max_{1\leq j\leq k}|\omega_i \cap  c_j|.
  $$
  

\begin{figure*}
{
\centering
\includegraphics[width=\linewidth]{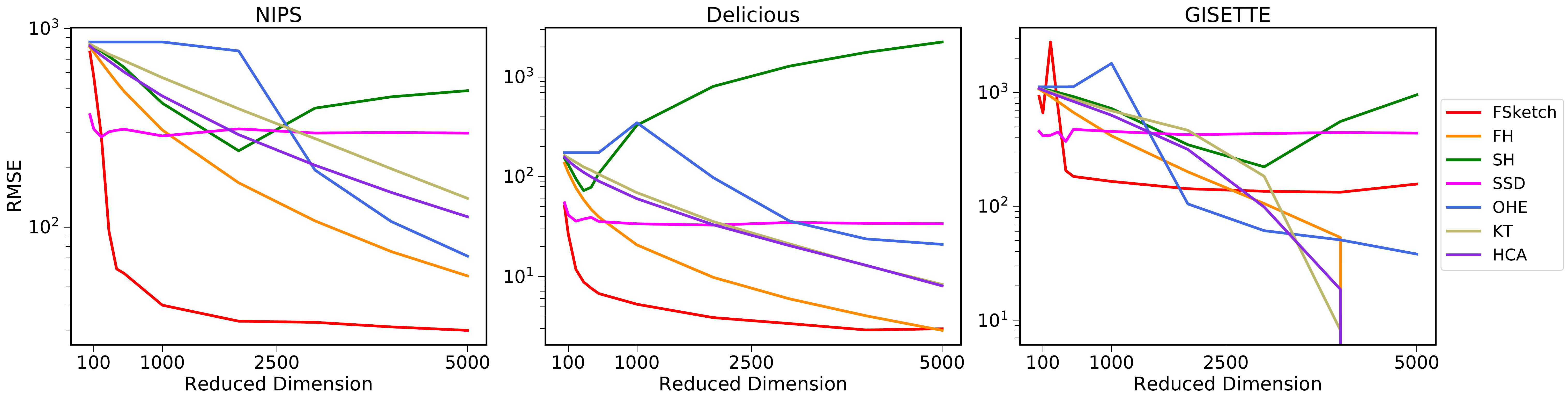}
    \caption{Comparison of $\RMSE$ values. A lower value is an indication of better performance. The GISETTE dataset is of 5000 dimensions and hence, FSketch suffers from an increase in RMSE as the embedding dimension also reaches 5000.}
\label{appendix:fig:rmse}
}
\end{figure*}

\begin{figure*}
{
\centering
    \includegraphics[width=0.75\linewidth]{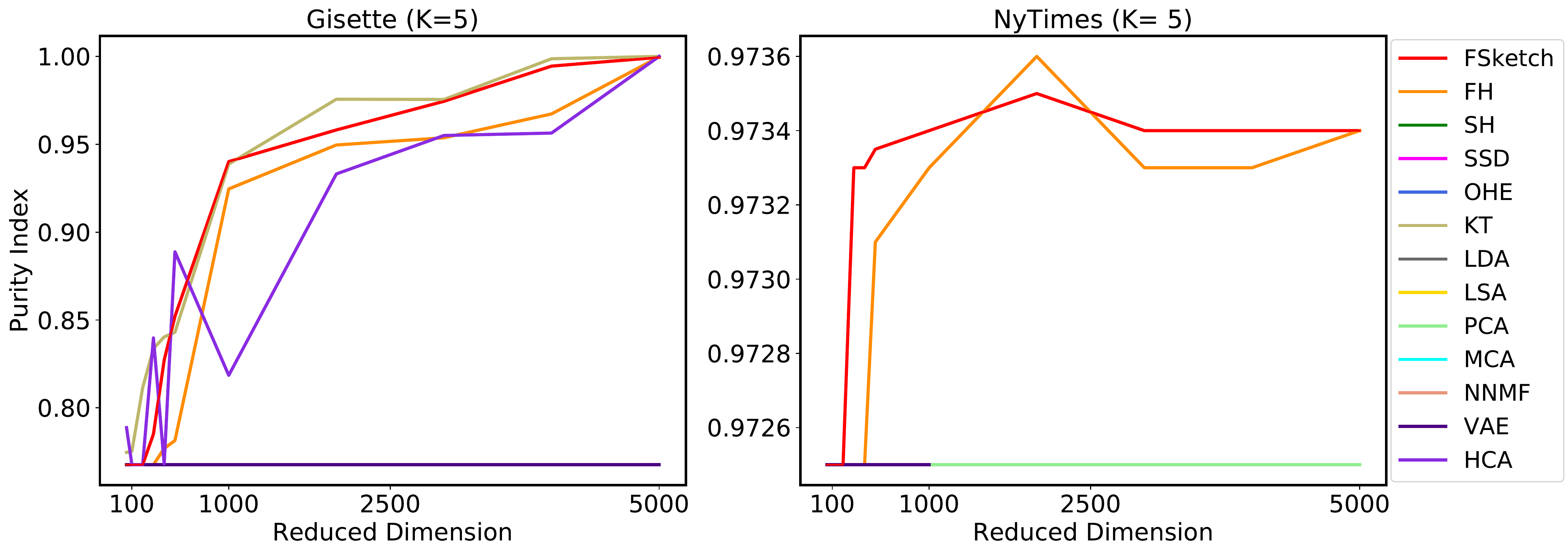}
\caption{{Comparing the quality of clusters on the compressed datasets. 
}}
\label{appendix:fig:purity}
}
\end{figure*}


\begin{figure*}
{
\centering
\includegraphics[width=\linewidth]{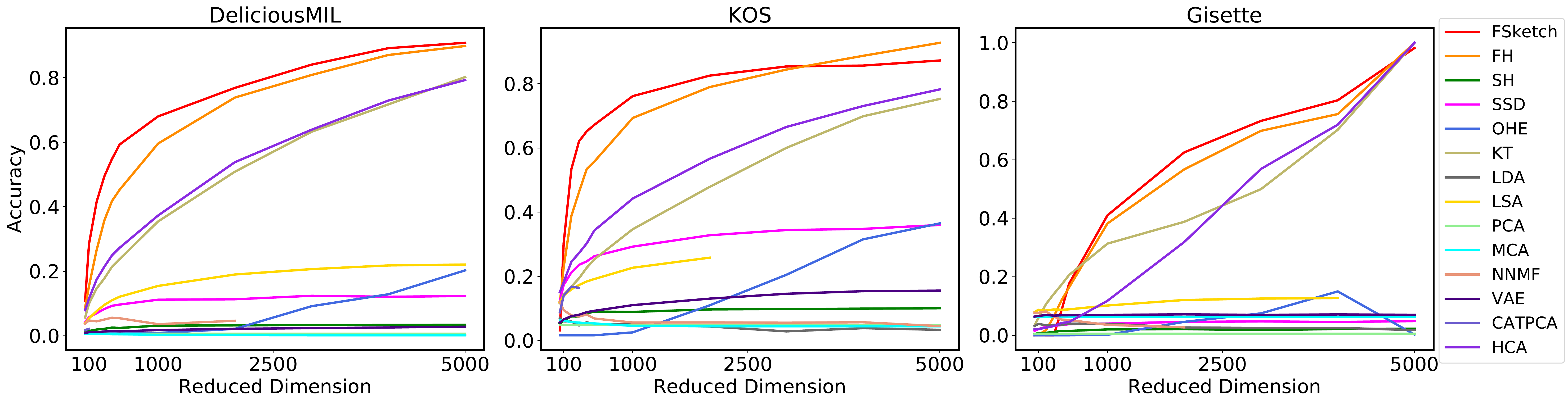}
\caption{{Comparing the performance of the similarity search task (estimating top-$k$ similar points with $k=100$) achieved on the reduced dimensional data obtained from various baselines.
}}
\label{appendix:fig:similarity_search}
}
\end{figure*}

\begin{figure*}
{
\centering
\includegraphics[width=\linewidth]{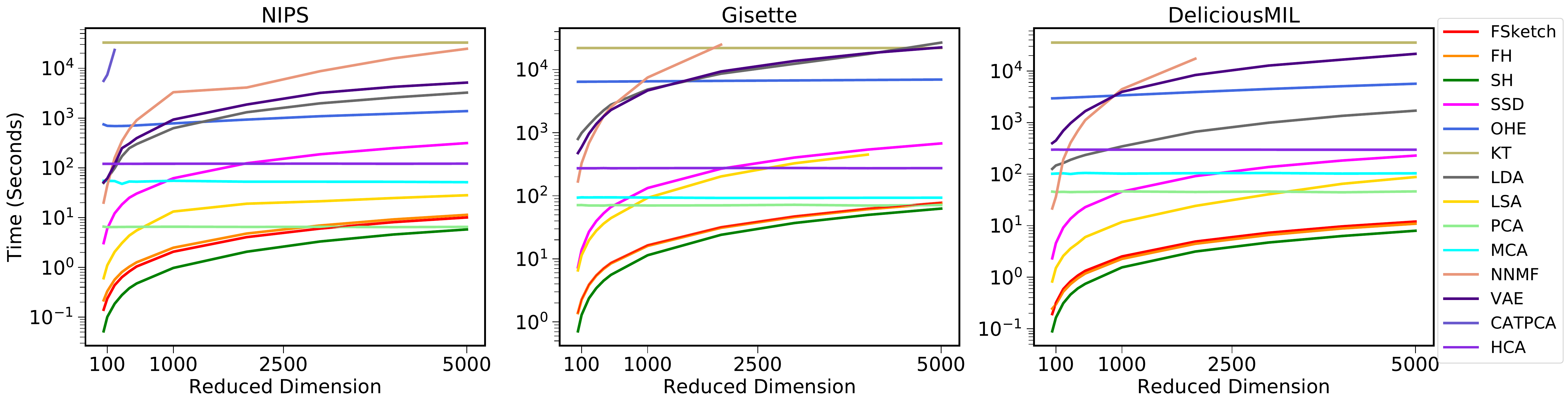}
\caption{Comparison of the dimensionality reduction times.}
\label{appendix:fig:reduction_time}
}
\end{figure*}

\section{Errors during dimensionality
reduction experiments}\label{sec:appendix_oom_error}
Several baselines give \textit{out-of-memory} error or their running time is quite high on some datasets. This makes it infeasible to include them in empirical comparison on $\RMSE$ and other end tasks.

We list these errors here. OHE gives \textit{out-of-memory} error for Brain cell dataset. { HCA gives DNS errors on NYTimes and BrainCell datasets. CATPCA could only on KOS and DeliciousMIL datasets that too upto only $300$ reduced dimension. Other than that it gives a DNS error.  VAE gives DNS errors on Enron datasets. }
 KT gives \textit{out-of-memory} error for NYTimes and Brain cell and on Enron it didn’t stop even after $10$ hrs. MCA also gives \textit{out-of-memory} error for NYTimes and Brain cell  datasets. Further, the dimensionality reduction time for NNMF was quite high -- on NYTimes it takes around $20$ hrs to do the dimensionality reduction for $3000$ dimension, and on the Brain cell dataset, NNMF didn't stop even after $10$ hrs. These errors prevented us from performing dimensionality reduction for all dimension using some of the algorithms.

\section{Extended experimental results}\label{appendix:section:extended_exp}
This section contains the remaining comparative plots for the RMSE (Figure~\ref{appendix:fig:rmse}), clustering (Figure~\ref{appendix:fig:purity}), similarity search experiments (Figure~\ref{appendix:fig:similarity_search}) and the dimensionality reduction time (Figure~\ref{appendix:fig:reduction_time}).

\section{\minfsketch: Combining multiple \fsketch}\label{subsec:minfsketch}

We proved in Lemma~\ref{lem:hconcentrationlemma} that our estimate $\hat{h}$ is within an additive error of $h$. 
A standard approach to improve the accuracy in such situations is to obtain several independent estimates and then compute a suitable statistic of the estimates. We were faced with a choice of mean, median and minimum of the estimates of which we decided to choose median after extensive empirical evaluation (see Section~{\ref{subsec:box_plot_median}}) and obtaining theoretical justification (explained in Section~\ref{appendix:subsec:minfsketch}). We first explain our algorithms in the next subsection.

\begin{figure}
    \centering
    \includegraphics[width=0.9\linewidth]{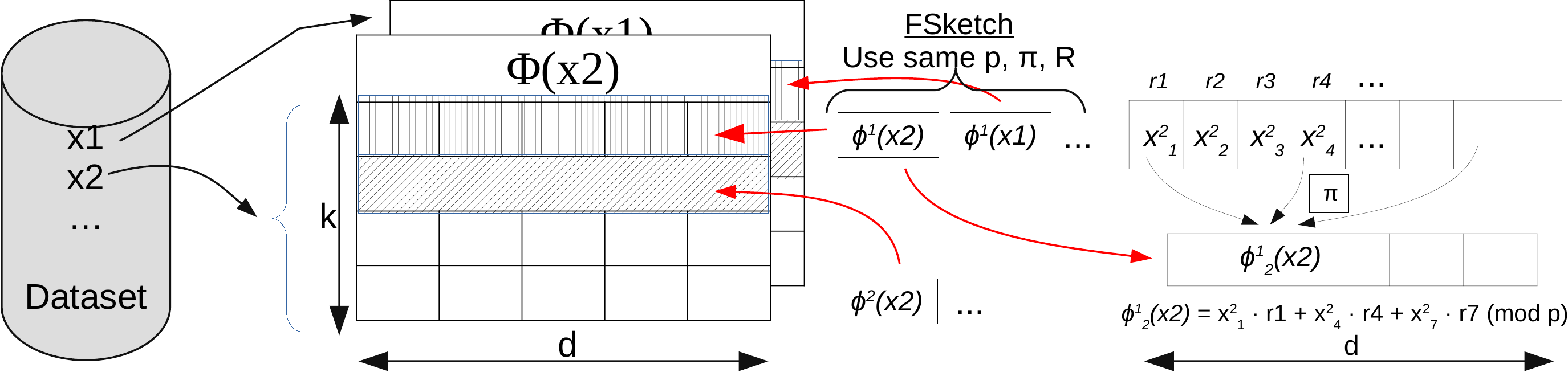}
    \caption{\minfsketch for categorical data --- sketch of each data point is a 2-dimensional array whose each row is an \fsketch. The $i$-th rows corresponding to all the data points use the same values of $\rho,R$.}
    \label{fig:catsketch-schema}
\end{figure}


\subsection{Algorithms for generating a sketch and estimaing Hamming distance}
Let $k,d$ be some suitably chosen integer parameters. An arity-$k$ dimension-$d$ \minfsketch for a categorical data, say $x$, is an array of $k$ sketches: $\Phi(x) = \langle \phi^1(x), \phi^2(x), \ldots \phi^k(x) \rangle$; the $i$-th entry of $\Phi(x)$ is a $d$-dimensional \fsketch. See Figure~\ref{fig:catsketch-schema} for an illustration. Note that the internal parameters $\rho,R,p$ required to run \fsketch to obtain the $i$-entry are same across all data points; the parameters corresponding to different $i$ are, however, chosen independently ($p$ can be the same).

Our algorithm for Hamming distance estimation is inspired from the Count-Median sketch~\cite{CORMODE200558} and Count sketch~\cite{CHARIKAR20043}. It estimates the Hamming distances between the pairs of ``rows'' from $\Phi(x)$ and $\Phi(y)$ and returns the median of the estimated distances. This procedure is followed in Algorithm~\ref{algo:fsketchestimate}.

\newcommand{\hxy}{\widehat{h}}

    \begin{algorithm}
	\noindent\hspace*{\algorithmicindent} \textbf{Input:} $\Phi(x)=\langle \phi^1(x), \phi^2(x), \ldots \phi^k(x) \rangle$, $\Phi(y)=\langle \phi^1(y), \phi^2(y), \ldots \phi^k(y) \rangle$
	\begin{algorithmic}[1]
	    \For{$i=1 \ldots k$}
            \State Compute $f=$ Hamming distance between $\phi^i(x)$ and $\phi^i(y)$
            \State If $f < dP$, $\hxy^i=\ln\left(1-\frac{f}{dP}\right)/\ln D$
            \State Else $\hxy^i=2\sigma$
        \EndFor
        \State \Return $\hat{h} = \min\{ \hxy^1, \hxy^2, \ldots \hxy^k\}$
	\end{algorithmic}
	\caption{Estimate Hamming distance between $x$ and $y$ from their \minfsketch\label{algo:fsketchestimate}}
    \end{algorithm}

\begin{figure*}[t]
\centering
\includegraphics[width=\linewidth]{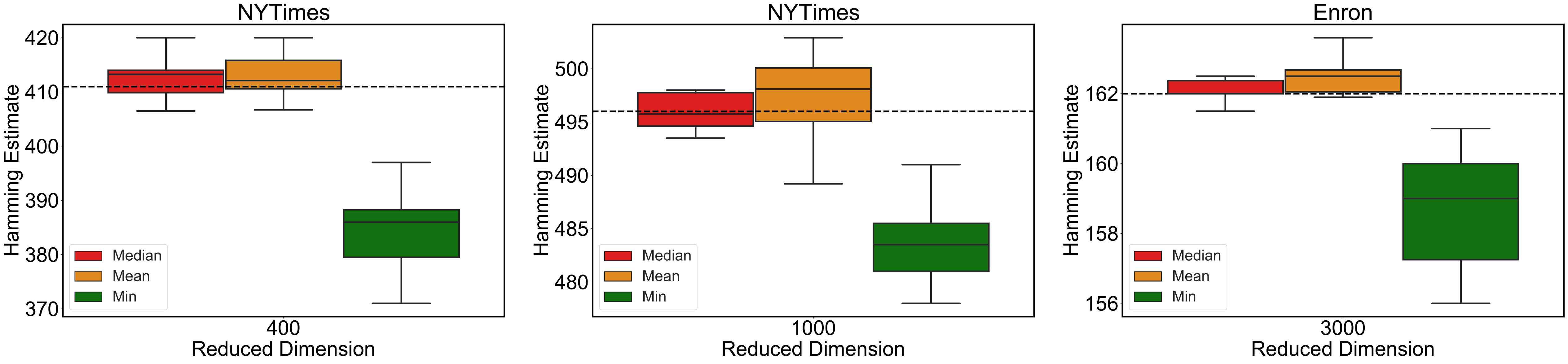}
\caption{{Box plot for the median, mean, and minimum of the \texttt{FSketch}’s estimate obtain \textit{via} it’s from its $10$ repetitions, then each experiment is repeated $10$ times for computing   the variance of these statistics.  The black dotted line corresponds to the actual Hamming distance. }}
\label{fig:box_plot_median_mean_min}
\end{figure*}

\subsection{Theoretical justification}\label{appendix:subsec:minfsketch}

%

We now give a proof that our \minfsketch estimator offers a better
approximation. Recall that $\sigma$ indicates the maximum number of non-zero
attributes in any data vector, and is often much small compared to the their
dimension, $n$. Surprisingly, our results are independent of $n$.
\begin{lem}\label{lem:medianlemma}
    Let $h^m$ denote the median of the estimates of Hamming distances obtained
    from $t$ independent
    \fsketch vectors of dimension $4\sigma$ and let $h$ denote the actual Hamming distance. Then,
    $$\Pr\big[|h^m - h| \ge 18\sqrt{\sigma}\big] \le \delta$$
    for any desired $\delta \in (0,1)$ if we use $t \ge 48\ln \tfrac{1}{\delta}$.
\end{lem}

\begin{proof}
    We start by using Lemma~\ref{lem:hconcentrationlemma} with
    $p=3$ and error ($\delta$ in the lemma statement) = $\tfrac{1}{4}$. Let
    $\hat{h}^i$ denote the $k$-th estimate. From the lemma we get that
    $$\Pr\big[ |\hat{h}^i - h| \ge 18\sqrt{\sigma} \big] \le \tfrac{1}{4}$$

    Define indicator random variables $W_1 \ldots W_t$ as $W_i=1$ iff
    $|\hat{h}^i - h| \ge 18\sqrt{\sigma}$. We immediately have $\Pr[W_i] \le
    \tfrac{1}{4}$. 
    Notice that $W_i=1$ can also be interpreted to indicate the event
    $h-18\sqrt{\sigma} \le \hat{h}^i \le h + 18 \sqrt{\sigma}$.
    Now, $h^m$ is the median of $\{\hat{h}^1, \hat{h}^2, \ldots
    \hat{h}^t\}$, and so, $h^m$ falls outside the range $[h-18\sqrt{\sigma},
    h+18\sqrt{\sigma}]$ only if more than half of the estimates fall outside this
    range., i.e., if $\sum_{i=1}^t W_i > t/2$. Since $\E[\sum_i W_i] \le t/4$, the probability of this event is
    easily bounded by $\exp{(-(\tfrac{1}{2}^2
    \tfrac{t}{4}/3))} = e^{-t/48} \le \delta$ using Chernoff's bound.
\end{proof}

\subsection{Choice of statistics in \minfsketch}\label{subsec:box_plot_median}
We conducted an experiment to decide whether to take median, mean or minimum of $k$ \fsketch estimates in the \minfsketch algorithm. We randomly sampled a pair of points and estimated the Hamming distance from its low-dimensional representation obtained from  \texttt{FSketch}. We repeated this $10$ times over different random mappings and computed the median, mean, and minimum of those $10$ different estimates. We further repeat this experiment $10$ times and generate a box-plot of the readings which is presented in Figure~\ref{fig:box_plot_median_mean_min}. We observe that median has the lowest variance and also closely estimates the actual Hamming distance between the pair of points.

\section{Dimensionality reduction algorithms}
\begin{table*}[h]
    \caption{A tabular summary of popular dimensionality reduction algorithms. Linear dimensionality reduction algorithms are those whose features in reduced dimension are linear combinations of the input features, and the others are known as non-linear algorithms. Supervised dimensionality reduction methods are those that require labelled datasets for dimensionality reduction. \label{table:dim-red}
}
  \begin{center}
        \begin{tabular} { | p {0.55cm} | p {2 cm} | p {4 cm} | p {2 cm} | p {2.5 cm} | p {2 cm} | p {2 cm} |}
        
        \hline
        S. No.
&Data type of input vectors &Objective/ Properties &Data type of sketch vectors &Result &Supervised or Unsupervised&Type of dimensionality reduction\\
\hline
1 &Real-valued vectors &Approximating pairwise euclidean distance, inner product
&Real-valued vectors&JL-lemma~\cite{JL83}&Unsupervised&Linear \\
\hline
2&Real-valued vectors&
Approximating pairwise euclidean distance, inner product
&Real-valued vectors&Feature Hashing~\cite{WeinbergerDLSA09}&Unsupervised&Linear \\
\hline
3&Real-valued vectors&Approximating pairwise cosine or angular similarity &Binary vectors&SimHash~\cite{simhash}&Unsupervised&Non-Linear\\
\hline
4&Real-valued vectors&Approximating pairwise $\ell_p$ norm for $p \in (0, 2]$&Real-valued vectors&$p$-stable random projection (SSD)~\cite{Indyk06}
&Unsupervised &Linear\\
\hline
5&Sets&Approximating pairwise Jaccard similarity &Integer valued vectors
&MinHash~\cite{BroderCFM98}&Unsupervised&Non-linear\\
\hline
6&Sparse binary vectors
&Approximating pairwise Hamming distance, Inner product, Jaccard and Cosine similarity 
&Binary vectors&BinSketch~\cite{ICDM}&Unsupervised&Non-linear\\
\hline
7&Real-valued vectors&Minimize  the variance in low dimension 
&Real-valued vectors&Principal Component Analysis (PCA)&Unsupervised
&Linear\\
\hline
8&Real-valued vectors (labelled input)&Maximizes class separability in the reduced dimensional space&Real-valued vectors&Linear Discriminant Analysis~\cite{FLDA} &Supervised&Linear\\
\hline
9&Real-valued vectors&
Embedding high-dimensional data for visualization in a low-dimensional space of two or three dimensions&
Real-valued vectors &$t$-SNE~\cite{vanDerMaaten2008}
&Unsupervised&Non-linear\\
\hline
10&Real-valued vectors &Minimize the reconstruction error 
&Real-valued vectors&Auto-encoder~\cite{10.5555/2987189.2987190}
&Unsupervised&Non-linear\\
\hline
11&Real-valued vectors
&Extracting nonlinear structures in low-dimension via Kernel function
&Real-valued vectors&Kernel-PCA~\cite{NIPS1998_226d1f15}&Unsupervised&Non-linear\\
\hline
 12 &Real-valued vectors &
Factorize input matrix into two small size non-negative matrices &Real-valued vectors &Non-negative matrix factorization (NNMF)~\cite{NNMF}
 &Unsupervised &Linear\\
\hline
13 &Real-valued
vectors &Compute a \textit{quasi-isometric}  low-dimensional embedding &Real-valued
vectors&Isomap~\cite{tenenbaum_global_2000}
&Unsupervised&Non-linear\\
\hline
14&
Real-valued
vectors
&Preserves the \textit{topological structure} of the data
&Real-valued
vectors
&Self-organizing map
~\cite{kohonen-self-organized-formation-1982}
&Unsupervised&Non-linear\\
\hline
\end{tabular}
    \end{center}
    \end{table*}

\end{document}